\documentclass[11pt]{article}

\usepackage[final]{acl}

\usepackage{times}
\usepackage{latexsym}

\usepackage[T1]{fontenc}

\usepackage[utf8]{inputenc}

\usepackage{microtype}

\usepackage{inconsolata}

\usepackage{graphicx}

\usepackage{subcaption}
\usepackage{booktabs} 

\usepackage{hyperref}
\usepackage{url}

\usepackage{amsmath}
\usepackage{amssymb}
\usepackage{mathtools}
\usepackage{amsthm}

\usepackage{algorithm}
\usepackage{algorithmic}

\usepackage{multirow}
\usepackage{enumitem}
\usepackage[dvipsnames, table]{xcolor}         
\usepackage{listings}
\usepackage{listingsutf8}
\usepackage[many]{tcolorbox}    	
\tcbuselibrary{listings, breakable}
\newtcolorbox{promptBox}[1]{
    title=\textbf{#1},
    breakable,
    fonttitle=\bfseries,
    boxrule = 1pt,
    toprule = 3pt, 
    colframe = RoyalBlue,
    enhanced,
    rounded corners,
    arc = 2pt,   
    top=1mm,bottom=1mm,left=1mm,right=1mm    
}

\theoremstyle{plain}

\theoremstyle{definition}

\theoremstyle{remark}

\usepackage{wrapfig}

\renewcommand{\algorithmiccomment}[1]{\textcolor{RoyalBlue}{$\triangleright$ #1}}

\newcommand{\ours}{TS-Debate}

\title{Multimodal Collaborative Debate for Zero-Shot Time Series Reasoning}

\author{
 \textbf{Patara Trirat\textsuperscript{1}}~
 \textbf{Jin Myung Kwak\textsuperscript{1,2}}~
 \textbf{Jay Heo\textsuperscript{1}}~
 \textbf{Heejun Lee\textsuperscript{1,2}}~
 \textbf{Sung Ju Hwang\textsuperscript{1,2}}
\\
 \textsuperscript{1}DeepAuto.ai~~\textsuperscript{2}KAIST
\\
 \small{
   \textbf{Correspondence:} \texttt{\{patara, sjhwang\}@deepauto.ai}
 }
}

\begin{document}
\maketitle
\begin{abstract}
Large language models\,(LLMs) are increasingly used as natural-language interfaces to structured data, yet they remain brittle when reasoning over time series. Visual patterns can be misleading, numerical claims can be hallucinated, and textual context can override evidence from the signal. We study \emph{zero-shot} time-series reasoning as a multimodal evidence arbitration problem for LLM agents. We propose \textbf{\ours{}}, an inference-time multi-agent protocol that requires no task-specific fine-tuning. \ours{} first elicits relevant domain knowledge, then assigns modality-specialized agents to textual context, visual patterns, and numerical signals, and coordinates their interaction through a verification-conflict-calibration procedure. Reviewer agents check decision-critical claims with lightweight code execution and numerical lookup, resolve cross-modal disagreement, and calibrate the final answer. Unlike generic multi-agent debate or unconstrained tool use, \ours{} specifies how evidence is exposed, which claims are checkable, and how verification outcomes shape synthesis. Across 20 tasks from three public benchmarks, \ours{} improves classification and question answering performance over strong baselines, while revealing that debate is most useful for global-structure and cross-view reasoning rather than local value reconstruction.
\end{abstract}


\section{Introduction} \label{section:introduction}

Time-series data underlie critical decisions in finance, climatology, and industrial monitoring \citep{trirat2024universal}. Beyond forecasting, practitioners increasingly ask \emph{reasoning-centric} questions such as: \emph{What explains this anomaly?}, \emph{Which historical pattern best matches this trajectory?}, or \emph{Will the underlying cause of this trend persist?} Answering such questions requires integrating numerical precision, temporal structure, and domain knowledge---tasks that remain challenging for current large language models\,(LLMs). As LLMs become a universal interface for data analysis, enabling robust \emph{time-series reasoning}\,(TSR) in natural language is both practically important and scientifically unresolved\,\citep{TSandLanguage_EMNLP_24, Reasoning_Agentic_TSLLM_25}.

Recent work suggests that TSR is inherently \emph{multimodal}\,\citep{EmpoweringTSR_MLLM_25, HowTSA_Multimodal_Survey_25}. Visual prompting compresses long numerical sequences into charts that multimodal LLMs can parse efficiently\,\citep{ByMyEyes_EMNLP_24}, and vision--language approaches show that time- and frequency-domain visualizations expose temporal structure that text-only prompting misses\,\citep{VLTime_NAACL_25}. However, simply providing multimodal inputs is insufficient. Without structured orchestration, additional modalities produce interference rather than improvement. Models over-trust visual scale impressions, hallucinate numeric details, or default to shallow linguistic heuristics when modalities disagree. Rather than asking whether multimodality helps TSR, we ask: \emph{\textbf{what inference-time protocol over modality-specialized agents is needed to convert multimodal access into reliable \underline{zero-shot} reasoning?}}

\begin{figure}[t]
    \centering
    \includegraphics[width=\linewidth]{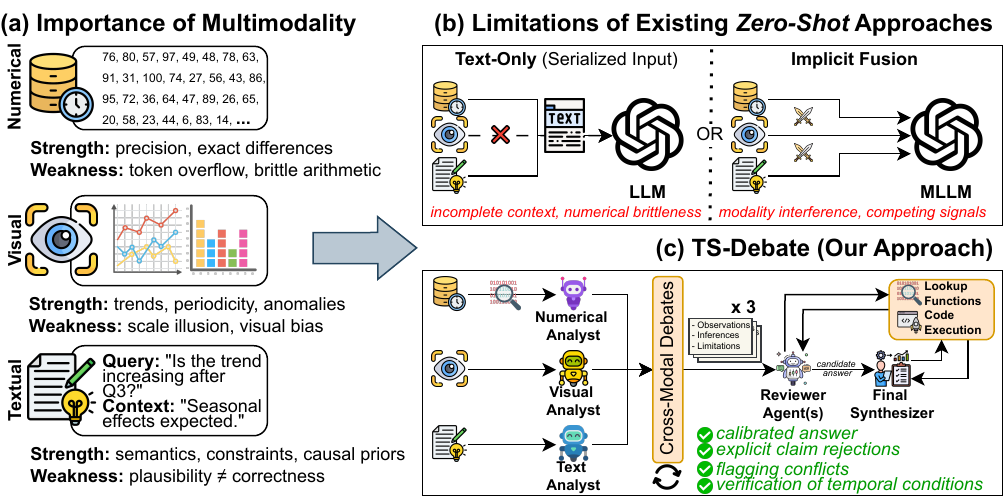}
    \caption{Overview of multimodal time-series reasoning: (a) complementary strengths and failure modes across modalities; (b) limitations of existing zero-shot approaches; and (c) \ours{}, a cross-modal agentic framework with explicit verification.} \label{figure:tsr_challenges}
    \vspace*{-0.5cm}
\end{figure}

Existing zero-shot approaches fall short not due to insufficient modality access or tooling, but because they lack a reasoning protocol tailored to the unique challenges of time-series data. As illustrated in \autoref{figure:tsr_challenges}, we identify three such challenges. \textbf{(i) Modality interference:} prior studies\,\citep{Reasoning_Agentic_TSLLM_25, MONAQ_EMNLP_25} either serialize all modalities into text---leading to context overflow and brittle arithmetic---or fuse them implicitly within a single model, leaving numeric, visual, and textual signals indistinguishable and impossible to selectively trust. \textbf{(ii) Numeric brittleness:} even when models identify the correct qualitative pattern, final answers frequently contain arithmetic errors or unjustified extrapolations, and classical debate frameworks\,\citep{MAD_EMNLP_24, MAD_Strategy_ICML_24} resolve disagreement by persuasion rather than verification---inadequate when claims about trends, extrema, and temporal conditions are objectively checkable. \textbf{(iii) Temporal-scope ambiguity:} verifiable past-present claims are routinely conflated with unverifiable future claims, so calibration cannot distinguish what the signal supports from what the model speculates.

To address these challenges, we introduce \emph{\textbf{\ours{}}}, a cross-modal, collaborative agentic \underline{debate} framework for zero-shot \underline{t}ime-\underline{s}eries reasoning. \ours{} instantiates modality-specialized agents---a text analyst, a visual analyst, and a numerical analyst---that produce explicitly labeled observations, inferences, and limitations grounded in their modality. A preliminary \emph{knowledge elicitation} stage activates relevant domain priors before any data analysis. Crucially, \ours{} replaces majority voting with a \emph{Verification-Conflict-Calibration}\,(VCC) protocol: reviewer agents verify temporally checkable claims via numerical lookup and code execution, type cross-modal conflicts, and calibrate answers based on both data consistency and temporal scope. A final synthesizer discounts---rather than implicitly fuses---unverified or contradicted claims. Our contribution is not the use of LLM agents, multi-agent debate, multimodal prompting, or code execution individually---each is established. Rather, we identify three challenges specific to zero-shot TSR and propose an inference-time protocol that addresses all three jointly. Our \textbf{main contributions} are as follows.
\begin{itemize}[leftmargin=9pt, nosep, noitemsep]
    \item We formalize TSR as a \emph{multimodal evidence arbitration} problem requiring modality-isolated evidence generation rather than implicit fusion.
    
    \item We propose \ours{}, an inference-time LLM-agent protocol that separates modality-specific evidence generation from claim verification and calibrated synthesis.
    
    \item We introduce the VCC protocol, which replaces persuasion- or vote-based debate resolution with programmatic verification, explicit conflict typing, and temporal-scope-aware calibration.

    \item We demonstrate consistent empirical improvements over competitive baselines across 20 tasks on three public benchmarks---MTBench (+7.39\%), TimerBed (+22.74\%), and TSQA (+21.58\%)---that are statistically significant in aggregate and persist under tool-matched controls, indicating that the gains stem from the protocol rather than from component access.
\end{itemize}
\begin{table*}[t]
\centering
\resizebox{\textwidth}{!}{%
\begin{tabular}{@{}l|ccccc|ccc@{}}
\toprule
\multirow{2}{*}{\textbf{Framework}} & \multicolumn{5}{c}{\textbf{Key Functionality}} & \multicolumn{3}{|c}{\textbf{Support Modality for Reasoning}} \\ \cmidrule(l){2-9} 
 & \textbf{Zero-Shot Reasoning} & \textbf{Multi-Agent Debate} & \textbf{Program-Aided Verification} & \textbf{Self-Elicited  Knowledge} & \textbf{Calibration Protocol} & \textbf{Raw Number} & \textbf{Textual Context} & \textbf{Visual Representation} \\ \midrule \midrule
TS-Reasoner\,\citep{TSReasoner_25} & $\textcolor{red}{\times}$ & $\textcolor{red}{\times}$ & $\textcolor{red}{\times}$ & $\textcolor{red}{\times}$ & $\textcolor{red}{\times}$ & $\textcolor{teal}{\checkmark}$ & $\textcolor{teal}{\checkmark}$ & $\textcolor{brown}{\triangle}$ (for training) \\
ChatTime\,\citep{ChatTime_AAAI_25} & $\textcolor{red}{\times}$ & $\textcolor{red}{\times}$ & $\textcolor{red}{\times}$ & $\textcolor{red}{\times}$ & $\textcolor{red}{\times}$ & $\textcolor{teal}{\checkmark}$ & $\textcolor{teal}{\checkmark}$ & $\textcolor{red}{\times}$ \\
Chat-TS\,\citep{Chat-TS_25} & $\textcolor{red}{\times}$ & $\textcolor{red}{\times}$ & $\textcolor{red}{\times}$ & $\textcolor{red}{\times}$ & $\textcolor{red}{\times}$ & $\textcolor{teal}{\checkmark}$ & $\textcolor{teal}{\checkmark}$ & $\textcolor{red}{\times}$ \\
ChatTS\,\citep{ChatTS_VLDB_25} & $\textcolor{red}{\times}$ & $\textcolor{red}{\times}$ & $\textcolor{red}{\times}$ & $\textcolor{red}{\times}$ & $\textcolor{red}{\times}$ & $\textcolor{teal}{\checkmark}$ & $\textcolor{teal}{\checkmark}$ & $\textcolor{red}{\times}$ \\
TimeMQA\,\citep{TimeMQA_ACL_25} & $\textcolor{red}{\times}$ & $\textcolor{red}{\times}$ & $\textcolor{red}{\times}$ & $\textcolor{red}{\times}$ & $\textcolor{red}{\times}$ & $\textcolor{teal}{\checkmark}$ & $\textcolor{teal}{\checkmark}$ & $\textcolor{red}{\times}$ \\
ITFormer\,\citep{ITFormer_ICML_25} & $\textcolor{red}{\times}$ & $\textcolor{red}{\times}$ & $\textcolor{red}{\times}$ & $\textcolor{red}{\times}$ & $\textcolor{red}{\times}$ & $\textcolor{brown}{\triangle}$ (for training) & $\textcolor{teal}{\checkmark}$ & $\textcolor{teal}{\checkmark}$ \\
TimeART\,\citep{TimeART_2026} & $\textcolor{red}{\times}$ & $\textcolor{red}{\times}$ & $\textcolor{teal}{\checkmark}$ & $\textcolor{red}{\times}$ & $\textcolor{red}{\times}$ & $\textcolor{teal}{\checkmark}$ & $\textcolor{teal}{\checkmark}$ & $\textcolor{red}{\times}$ \\
VL-Time\,\citep{VLTime_NAACL_25} & $\textcolor{teal}{\checkmark}$ & $\textcolor{red}{\times}$ & $\textcolor{red}{\times}$ & $\textcolor{red}{\times}$ & $\textcolor{red}{\times}$ & $\textcolor{red}{\times}$ & $\textcolor{teal}{\checkmark}$ & $\textcolor{teal}{\checkmark}$ \\
ByMyEyes\,\citep{ByMyEyes_EMNLP_24} & $\textcolor{teal}{\checkmark}$ & $\textcolor{red}{\times}$ & $\textcolor{red}{\times}$ & $\textcolor{red}{\times}$ & $\textcolor{red}{\times}$ & $\textcolor{red}{\times}$ & $\textcolor{teal}{\checkmark}$ & $\textcolor{teal}{\checkmark}$ \\ \midrule
\emph{\textbf{\ours{}}} (Ours) & $\textcolor{teal}{\checkmark}$ & $\textcolor{teal}{\checkmark}$ & $\textcolor{teal}{\checkmark}$ & $\textcolor{teal}{\checkmark}$ & $\textcolor{teal}{\checkmark}$ & $\textcolor{teal}{\checkmark}$ & $\textcolor{teal}{\checkmark}$ & $\textcolor{teal}{\checkmark}$ \\ \bottomrule
\end{tabular}%
}
\caption{Comparison between \ours{} and existing time-series reasoning frameworks.} \label{table:framework_comparison}
\vspace*{-0.2cm}
\end{table*}
\section{Related Work} \label{section:related_work}

\paragraph{Time-Series Analysis and Reasoning} LLMs have been applied to TSR via numerical serialization\,\citep{LLMTime_NeurIPS_23}, alignment-based adapters\,\citep{TimeLLM_ICLR_24,toward_tsr_NeurIPSW_24}, and coupling with specialized encoders\,\citep{TSReasoner_25,ChatTime_AAAI_25}. These methods focus on representation learning and typically operate within a single numerical channel, requiring task-specific training. In contrast, \ours{} assumes off-the-shelf LLM and MLLM backbones and targets an \emph{inference-time orchestration protocol}, coordinating modality-specialized agents under explicit verification rather than training new encoders.

\paragraph{Multimodal LLMs for Time Series} A parallel line converts time series into visual representations for MLLM reasoning, including waveforms, spectrograms\,\citep{VLTime_NAACL_25}, and sensor plots\,\citep{ByMyEyes_EMNLP_24,EmpoweringTSR_MLLM_25}, achieving gains with reduced token budgets. However, these approaches rely on a \emph{single fusion step} within an MLLM. \ours{} instead instantiates modality-isolated agents for numerical, visual, and textual evidence and coordinates them through a multi-stage protocol, allowing cross-modal evidence to interact iteratively rather than implicitly fusing them.

\paragraph{Multi-Agent Debate} Multi-agent debate\,(MAD) frameworks\,\citep{MAD_Strategy_ICML_24,SparseMAD_EMNLP_24,Vote_vs_Consensus_ACL_25,StopOvervaluing_25} use LLM agents that critique one another to improve reasoning, but are predominantly text-only and assume homogeneous agents with shared inputs. \ours{} differs in both assumptions and architecture: agents are \emph{modality-isolated} and \emph{collaborative} rather than adversarial, and judges are equipped with code execution to verify numerical and temporal claims. Accordingly, interaction in \ours{} is not adversarial debate for answer selection, but an inference-time protocol for reconciling complementary multimodal evidence. To our knowledge, neither prior MAD frameworks nor agentic or visual TSR systems jointly support modality-isolated evidence generation, programmatic claim verification, explicit conflict typing, and temporal-scope-aware calibration\,(\autoref{table:framework_comparison}).


\section{Proposed Framework: \emph{\ours{}}} \label{section:method}

\subsection{Problem Setting} \label{section:problem_setting}

We study \emph{zero-shot} TSR over univariate ($d{=}1$) or multivariate ($d{>}1$) sequences. Each instance consists of a natural-language task specification $q$, an observed sequence $\mathbf{x}_{1:T} \in \mathbb{R}^{T \times d}$, and auxiliary context $c$\,(e.g., textual descriptions or metadata). The goal is to produce an answer $\hat{a} \in \mathcal{A}(q)$, where the answer space is task-induced\,(e.g., a class label, a numerical vector, or a multiple-choice option). \ours{} performs no task-specific training and does not update parameters at inference time; instead, it orchestrates LLM agents, each operating on a different representational view of the same signal, to construct an evidence-grounded response.

To account for the temporal nature of time series, we distinguish \emph{past-present} tasks, whose target lies within the observed horizon, from \emph{future} tasks, whose target lies beyond it. This distinction determines verification semantics: past-present claims can be checked directly against $\mathbf{x}_{1:T}$, whereas future claims must be evaluated through domain-consistent reasoning and uncertainty calibration.

\begin{figure*}[t]
    \centering
    \includegraphics[width=\linewidth]{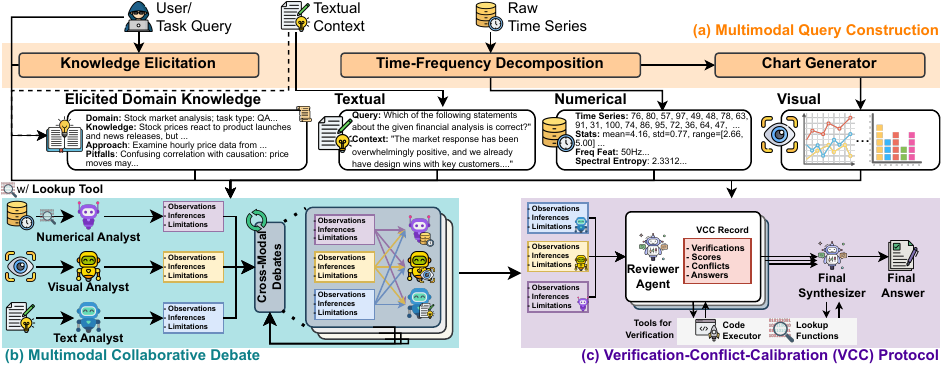}
    \vspace*{-0.7cm}
    \caption{Overview of the \ours{} framework. \textbf{(a) Multimodal query construction} derives textual, numerical, and visual representations from user queries, raw time series, and contextual information. \textbf{(b) Multimodal collaborative debate} enables numerical, visual, and text analysts to reason over these representations and produce structured observations, inferences, and limitations. \textbf{(c) VCC protocol} applies tool-assisted review to assess debate outputs, identify conflicts, and synthesize the final answer.} \label{figure:framework_overview}
    \vspace*{-0.5cm}
\end{figure*}

\subsection{Framework Overview} \label{section:overview}

Let $\mathcal{M}=\{\textsf{T},\textsf{V},\textsf{N}\}$ denote text, visual, and numerical modalities. As illustrated in \autoref{figure:framework_overview}\,(and Algorithm~\ref{alg:pseudo_code}), \ours{} proceeds in three stages. \underline{First}, it elicits task- and domain-specific priors from $q$\,(optionally with $c$) to form shared guidance for all agents. \underline{Second}, it constructs modality-specific interfaces---textual context for semantics, charts for global structure, and precise numerical evidence---and runs an iterative collaborative debate among modality-specialized analysts. \underline{Third}, it applies a verification-conflict-calibration protocol: reviewer agents verify analysts' claims, explicitly identify and resolve conflicts when possible, and produce calibrated candidate answers, after which a final synthesizer evaluates reviewer reasoning quality and produces the final answer $\hat{a}$.

\subsection{Multimodal Query Construction} \label{section:mqc}

A central design principle of \ours{} is to expose \emph{complementary and verifiable} views of $\mathbf{x}_{1:T}$ to different agents, reducing reliance on any single representation. This is implemented via task-conditioned knowledge elicitation, multiple representational views of the signal, and explicit visual and numerical query interfaces.

\paragraph{Knowledge Elicitation} Before interpreting any data, \ours{} elicits domain priors $k=\mathsf{Elicit}(q, c)$, which serve as a \emph{shared analysis contract} for all agents. Concretely, $k$ encodes a characterization of the task/domain, domain constraints and expected patterns, salient signal attributes, a recommended analysis procedure, common failure modes, and guidance on which modalities are most informative. This anchors zero-shot reasoning to task semantics and domain methodology before any modality-specific analysis.

\paragraph{Time-Frequency Views} From $\mathbf{x}_{1:T}$, we derive complementary representational views below.
\begin{itemize}[leftmargin=9pt, nosep, noitemsep]
    \item \textbf{Temporal View}\,($v_{\textrm{time}}$): a time-domain representation emphasizing trends, change points, and extrema.
    \item \textbf{Spectral View}\,($v_{\textrm{freq}}$): a frequency-domain representation highlighting periodicities and oscillatory structure\,(e.g., seasonality and recurrences).
    \item \textbf{Feature/Statistical View}\,($v_{\textrm{feat}}$): summary statistics and detected events\,(e.g., peaks, troughs, and anomalies) for precise quantitative checks.
\end{itemize}

\paragraph{Multimodal Query Interfaces} \ours{} exposes these views through two explicit interfaces.
\begin{itemize}[leftmargin=9pt, nosep, noitemsep]
    \item \textbf{Visual Interface.} A time-domain chart $\mathsf{Chart}_{\textrm{time}}$ and a frequency-domain chart $\mathsf{Chart}_{\textrm{freq}}$ form the input substrate for the visual analyst, surfacing global structure\,(e.g., trend, volatility regimes, and periodicity) that is difficult to infer from numerical probes.
    \item \textbf{Numerical Interface.} Controlled tool access to the time series and derived attributes, denoted $\textsf{Lookup}(\cdot)$, supports querying values over intervals, probing neighborhoods, retrieving detected events/features, and obtaining frequency-domain information. This interface is essential for \emph{verifiability}, grounding claims in concrete values.
\end{itemize}

\subsection{Multimodal Collaborative Debate} \label{section:main_debate}
\ours{} uses \emph{collaborative debate} to elicit diverse evidence without adversarial contest, refining hypotheses across modalities while explicitly acknowledging uncertainty and limitations.

\paragraph{Modality-Specialized Analysts} For each modality $m \in \mathcal{M}$, \ours{} instantiates an analyst $\mathcal{A}_m$.
\begin{itemize}[leftmargin=9pt, nosep, noitemsep]
    \item \textbf{Text analyst}\,$\mathcal{A}_{\textsf{T}}(q,c,k)$ interprets contextual constraints and textual descriptions, without direct numerical verification.
    \item \textbf{Visual analyst}\,$\mathcal{A}_{\textsf{V}}(q,\mathsf{Chart}_{\textrm{time}},\mathsf{Chart}_{\textrm{freq}},k)$ extracts qualitative, global structure from the generated charts.
    \item \textbf{Numerical analyst}\,$\mathcal{A}_{\textsf{N}}(q,\mathsf{Lookup}(\mathbf{x}_{1:T}),k)$ produces precise, checkable statements about values and derived features via numerical queries.

\end{itemize}
All analysts share the same $k$, so disagreements reflect evidence and representational differences rather than inconsistent task interpretations.

\paragraph{Structured Evidence Generation} Each analyst produces an evidence report $e_m^{(r)}$ at round $r$ in a template that separates \emph{verifiable observations} from \emph{interpretive inferences}, comprising
\begin{itemize}[leftmargin=9pt, nosep, noitemsep]
    \item \textbf{Understanding}: a restatement of the task $q$ in the analyst's own words.
    \item \textbf{Observations}: modality-grounded statements that are, in principle, checkable against the provided inputs\,(text, charts, or numbers), explicitly labeled as observations.
    \item \textbf{Inferences}: conclusions derived from observations using $k$, explicitly labeled as inferences.
    \item \textbf{Limits}: what cannot be determined reliably from the analyst's modality and inputs.
\end{itemize}

\paragraph{Iterative Evidence Refinement} Debate proceeds for $R \ge 1$ rounds. In round $1$, analysts work \emph{independently} to maximize epistemic diversity. Thus, 
\begin{equation} \label{eq:debate_first}
    e_m^{(1)} = \mathcal{A}_m(q,I_m,k), \quad m\in\mathcal{M},
\end{equation}
where $I_{\textsf{T}} = c$, $I_{\textsf{V}} = [\mathsf{Chart}_{\textrm{time}},\mathsf{Chart}_{\textrm{freq}}]$, and $I_{\textsf{N}}=\textsf{Lookup}(\mathbf{x}_{1:T})$. For rounds $r\ge2$, each analyst conditions on the others' previous evidence and may refine, correct, or strengthen its position:
\begin{equation} \label{eq:debate_rounds}
e_m^{(r)} = \mathcal{A}_m(q, I_m, k, \{e_{m'}^{(r-1)}\}_{m'\in\mathcal{M}}).
\end{equation}
Modality isolation thus constrains only first-round evidence generation. From round two onward, cross-modal integration is explicit and iterative, and reviewer agents subsequently verify each claim against \emph{all} modalities. Refinement is \emph{not} forced convergence. Analysts incorporate complementary observations while retaining principled disagreement when warranted by modality-specific evidence\,(e.g., a visually plausible pattern can be numerically refuted).

\subsection{Verification-Conflict-Calibration (VCC)} \label{section:vcc_protocol}

Evidence generation alone is insufficient: multimodal debate surfaces diverse hypotheses, but without disciplined validation and conflict handling, diversity can amplify error. VCC enforces explicit verification, conflict modeling, and calibrated synthesis, with the design intent that additional verified evidence monotonically expands error detection and correction, while unresolved conflicts trigger conservative calibration rather than confident error.

\paragraph{Claim Verification \& Domain Consistency} Let $\Omega$ be the set of atomic claims extracted from the final-round evidence $E^{(R)}=\{e_m^{(R)}\}_{m\in\mathcal{M}}$, where each $\omega\in\Omega$ is a minimal evaluable statement\,(e.g., ``the series increases sharply near $t\!=\!t_0$'' or ``a dominant period is present'').

For each claim $\omega$, a reviewer $\mathcal{A}_j$ assigns a verification status $V_j(\omega) \in \{V_\textsf{VERIFIED},V_\textsf{UNVERIFIED},V_\textsf{CONTRADICTED}\}$, determined by checking $\omega$ against available sources: \textbf{numerical} evidence via $\textsf{Lookup}(\cdot)$ and derived features; \textbf{computational} evidence via a lightweight code executor\,(e.g., arithmetic checks and aggregations); \textbf{visual} evidence via chart-level consistency checks; and \textbf{textual} evidence $(q,c)$ for task semantics and contextual constraints. Here, $V_\textsf{VERIFIED}$ indicates direct support, $V_\textsf{CONTRADICTED}$ direct refutation, and $V_\textsf{UNVERIFIED}$ insufficient evidence for either.

The reviewer also assigns a domain-consistency label $D_j(\omega) \in \{D_\textsf{MATCHES},D_\textsf{VIOLATES},D_\textsf{N/A}\}$, evaluating whether $\omega$ is consistent with constraints encoded in $k$ (with $D_\textsf{N/A}$ when domain constraints do not apply). The combined annotation $[V_j(\omega),D_j(\omega)]$ distinguishes unsupported claims from domain-inconsistent claims---critical when direct verification is unavailable\,(e.g., future targets) but domain constraints still apply.

Beyond claim-level labels, each $\mathcal{A}_j$ scores each analyst's evidence along three axes: $s_{j,m} = (s^{\textsf{inf}}_{j,m},\, s^{\textsf{obs}}_{j,m},\, s^{\textsf{hon}}_{j,m})$, with $s^{\textsf{inf}}_{j,m}\in[0,50]$, $s^{\textsf{obs}}_{j,m}\in[0,30]$, and $s^{\textsf{hon}}_{j,m}\in[0,20]$. These correspond to \textbf{inference quality}\,(logical soundness and calibration), \textbf{observation quality}\,(specificity, verifiability, and relevance), and \textbf{honesty}\,(explicit acknowledgment of limits and avoidance of unwarranted certainty). Reviewers are instructed to identify limitations even in strong responses, discouraging score saturation; scores guide synthesis toward evidence that is both grounded and appropriately uncertain.

\paragraph{Explicit Conflict Modeling} After verification, reviewers indicate outstanding conflicts via
$\Delta_j \in$ $\{\Delta_\textsf{NO\_CONFLICT}$,$\Delta_\textsf{DETECTED}$,$\Delta_\textsf{RESOLVED}\}$, where $\Delta_\textsf{DETECTED}$ flags incompatible claims with no verified resolution, $\Delta_\textsf{RESOLVED}$ marks conflicts resolved via verification and domain constraints, and $\Delta_\textsf{NO\_CONFLICT}$ indicates mutually consistent (or immaterial) evidence. 

\emph{Why explicit conflicts matter}. Averaging or voting over conflicting rationales is brittle. Modalities can disagree for principled reasons\,(e.g., a trend appears visually but disappears under numerical scrutiny), and a ``majority'' can be wrong if it shares a methodological flaw. VCC therefore handles conflicts explicitly so that disagreements trigger additional verification or conservative calibration rather than being smoothed away.

At the synthesizer level, disagreement is summarized in two ways. First, the \textbf{answer agreement pattern} $\Gamma \in$ $\{\Gamma_\textsf{UNANIMOUS}$, $\Gamma_\textsf{SPLIT}$, $\Gamma_\textsf{ALL\_DIFFERENT}\}$ captures whether reviewers converge. Second, the \textbf{resolution type} $\Lambda \in$ $\{\Lambda_\textsf{VERIFIED\_RESOLUTION}$, $\Lambda_\textsf{UNRESOLVED}$, $\Lambda_\textsf{NO\_CONFLICT}$, $\Lambda_\textsf{APPROACH\_ERROR}\}$ indicates whether disagreements were settled by verified evidence, remain unresolved, are absent, or stem from an incorrect approach relative to $k$, separating genuine ambiguity from methodological error.

\paragraph{Calibrated Answer Synthesis} Each reviewer $j$ produces a calibrated answer $\hat{a}_j$ and an accompanying VCC record
\begin{equation} \label{eq:vcc_output}
\begin{aligned}
    u_j \;=\; & (\hat{a}_j,\ \{(V_j(\omega),D_j(\omega))\}_{\omega\in\Omega},\\ &\Delta_j,\ \{s_{j,m}\}_{m\in\mathcal{M}}).
\end{aligned}
\end{equation}
Calibration follows explicit rules:
\begin{itemize}[leftmargin=9pt, nosep, noitemsep]
    \item If the key claims supporting $\hat{a}_j$ are $V_\textsf{VERIFIED}$ and $D_\textsf{MATCHES}$, and $\Delta_j=\Delta_\textsf{NO\_CONFLICT}$, the answer may be stated confidently.
    \item If key claims are $V_\textsf{UNVERIFIED}$ or $\Delta_j=\Delta_\textsf{DETECTED}$, the answer must be conservative and explicitly reflect uncertainty.
    \item If a claim is $D_\textsf{VIOLATES}$, it is rejected even if plausible from a single modality.
\end{itemize}
Temporal scope further modulates calibration. For \emph{past-present} tasks, direct verification against $\mathbf{x}_{1:T}$ dominates. For \emph{future-oriented} tasks, \ours{} cannot verify future outcomes from past data; instead, it evaluates whether the causes of observed patterns plausibly persist under $k$. This discourages naive extrapolation\,(``the trend continues because it has continued'') and promotes conditional predictions grounded in domain plausibility.

Finally, the synthesizer produces $\hat{a}$ and is explicitly \emph{not} a vote counter. It evaluates reviewers' \emph{reasoning quality} from their VCC records $u_j$ via: an \textbf{approach check} for adherence to the analysis procedure implied by $k$ and temporal scope; \textbf{reviewer scoring} on a fixed rubric\,(each criterion $0$--$20$) covering task understanding, evidence usage, verification discipline, conflict handling, and calibration; and \textbf{decision synthesis} that prioritizes answers supported by verified evidence and correct methodology. If all reviewers exhibit approach errors ($\Lambda_\textsf{APPROACH\_ERROR}$), the synthesizer re-verifies the most decision-critical claims and derives an answer directly. All prompts are provided in \S\ref{section:prompts}.

\section{Experiments} \label{section:experiments}
To verify the effectiveness of \ours{}, we conduct extensive experiments on a diverse set of TSR tasks spanning classification, regression, and question answering\,(QA). Additionally, we perform ablation and sensitivity analyses.

\begin{table*}[t]
\small
\centering
\resizebox{\textwidth}{!}{%
\begin{tabular}{@{}lccccccccc@{}}
\toprule
\textbf{Benchmarks} & \textbf{Zero-Shot} & \textbf{+ MM} & \textbf{CoT} & \textbf{+ MM} & \textbf{MAD} & \textbf{+ MM} & \textbf{ByMyEyes} & \textbf{VL-Time} & \textbf{\ours{}} \\ 
\midrule
\midrule
\multicolumn{10}{c}{\emph{Classification (Accuracy) - Higher is Better}} \\
MTBench (Finance)   & \begin{tabular}[c]{@{}c@{}}0.272\\ \small{($\pm$0.019)}\end{tabular}   & \begin{tabular}[c]{@{}c@{}}0.375\\ \small{($\pm$0.032)}\end{tabular}   & \begin{tabular}[c]{@{}c@{}}0.275\\ \small{($\pm$0.015)}\end{tabular}   & \begin{tabular}[c]{@{}c@{}}0.370\\ \small{($\pm$0.017)}\end{tabular}          & \begin{tabular}[c]{@{}c@{}}0.300\\ \small{($\pm$0.023)}\end{tabular}   & \begin{tabular}[c]{@{}c@{}}0.347\\ \small{($\pm$0.027)}\end{tabular}   & \begin{tabular}[c]{@{}c@{}}0.445\\ \small{($\pm$0.018)}\end{tabular}     & \begin{tabular}[c]{@{}c@{}}0.300\\ \small{($\pm$0.025)}\end{tabular}   & \textbf{\begin{tabular}[c]{@{}c@{}}0.543$^{\dagger}$\\ \small{($\pm$0.043)}\end{tabular}}  \\
MTBench (Weather)   & \begin{tabular}[c]{@{}c@{}}0.250\\ \small{($\pm$0.008)}\end{tabular}   & \begin{tabular}[c]{@{}c@{}}0.523\\ \small{($\pm$0.021)}\end{tabular}   & \begin{tabular}[c]{@{}c@{}}0.220\\ \small{($\pm$0.036)}\end{tabular}   & \begin{tabular}[c]{@{}c@{}}0.477\\ \small{($\pm$0.039)}\end{tabular}          & \begin{tabular}[c]{@{}c@{}}0.200\\ \small{($\pm$0.014)}\end{tabular}   & \begin{tabular}[c]{@{}c@{}}0.320\\ \small{($\pm$0.029)}\end{tabular}   & \begin{tabular}[c]{@{}c@{}}0.190\\ \small{($\pm$0.008)}\end{tabular}     & \begin{tabular}[c]{@{}c@{}}0.220\\ \small{($\pm$0.029)}\end{tabular}   & \textbf{\begin{tabular}[c]{@{}c@{}}0.557\\ \small{($\pm$0.025)}\end{tabular}}  \\
TimerBed            & \begin{tabular}[c]{@{}c@{}}0.286\\ \small{($\pm$0.017)}\end{tabular}   & \begin{tabular}[c]{@{}c@{}}0.317\\ \small{($\pm$0.018)}\end{tabular}   & \begin{tabular}[c]{@{}c@{}}0.343\\ \small{($\pm$0.020)}\end{tabular}   & \begin{tabular}[c]{@{}c@{}}0.332\\ \small{($\pm$0.023)}\end{tabular}          & \begin{tabular}[c]{@{}c@{}}0.294\\ \small{($\pm$0.014)}\end{tabular}   & \begin{tabular}[c]{@{}c@{}}0.317\\ \small{($\pm$0.024)}\end{tabular}   & \begin{tabular}[c]{@{}c@{}}0.290\\ \small{($\pm$0.032)}\end{tabular}     & \begin{tabular}[c]{@{}c@{}}0.332\\ \small{($\pm$0.024)}\end{tabular}   & \textbf{\begin{tabular}[c]{@{}c@{}}0.421$^{\dagger}$\\ \small{($\pm$0.051)}\end{tabular}}  \\
TSQA                & \begin{tabular}[c]{@{}c@{}}0.378\\ \small{($\pm$0.045)}\end{tabular}   & \begin{tabular}[c]{@{}c@{}}0.413\\ \small{($\pm$0.019)}\end{tabular}   & \begin{tabular}[c]{@{}c@{}}0.545\\ \small{($\pm$0.034)}\end{tabular}   & \begin{tabular}[c]{@{}c@{}}0.523\\ \small{($\pm$0.041)}\end{tabular}          & \begin{tabular}[c]{@{}c@{}}0.565\\ \small{($\pm$0.041)}\end{tabular}   & \begin{tabular}[c]{@{}c@{}}0.542\\ \small{($\pm$0.042)}\end{tabular}   & \begin{tabular}[c]{@{}c@{}}0.383\\ \small{($\pm$0.066)}\end{tabular}     & \begin{tabular}[c]{@{}c@{}}0.498\\ \small{($\pm$0.049)}\end{tabular}   & \textbf{\begin{tabular}[c]{@{}c@{}}0.615$^{\dagger}$\\ \small{($\pm$0.023)}\end{tabular}}  \\

\emph{\textbf{Average}}             & \begin{tabular}[c]{@{}c@{}}0.296\\ \small{($\pm$0.022)}\end{tabular}   & \begin{tabular}[c]{@{}c@{}}0.407\\ \small{($\pm$0.022)}\end{tabular}   & \begin{tabular}[c]{@{}c@{}}0.346\\ \small{($\pm$0.026)}\end{tabular}   & \begin{tabular}[c]{@{}c@{}}0.425\\ \small{($\pm$0.030)}\end{tabular}          & \begin{tabular}[c]{@{}c@{}}0.340\\ \small{($\pm$0.023)}\end{tabular}   & \begin{tabular}[c]{@{}c@{}}0.381\\ \small{($\pm$0.031)}\end{tabular}   & \begin{tabular}[c]{@{}c@{}}0.327\\ \small{($\pm$0.031)}\end{tabular}     & \begin{tabular}[c]{@{}c@{}}0.338\\ \small{($\pm$0.032)}\end{tabular}   & \textbf{\begin{tabular}[c]{@{}c@{}}0.534$^{\dagger}$\\ \small{($\pm$0.035)}\end{tabular}}  \\
\midrule
\multicolumn{10}{c}{\emph{Regression (MAE) - Lower is Better}} \\
MTBench (Finance)   & \begin{tabular}[c]{@{}c@{}}1.010\\ \small{($\pm$0.225)}\end{tabular}   & \begin{tabular}[c]{@{}c@{}}0.973\\ \small{($\pm$0.150)}\end{tabular}   & \begin{tabular}[c]{@{}c@{}}1.002\\ \small{($\pm$0.129)}\end{tabular}   & \begin{tabular}[c]{@{}c@{}}2.796\\ \small{($\pm$1.675)}\end{tabular}          & \begin{tabular}[c]{@{}c@{}}1.079\\ \small{($\pm$0.092)}\end{tabular}   & \begin{tabular}[c]{@{}c@{}}4.051\\ \small{($\pm$1.547)}\end{tabular}   & \begin{tabular}[c]{@{}c@{}}36.295\\ \small{($\pm$5.221)}\end{tabular}    & \begin{tabular}[c]{@{}c@{}}1.574\\ \small{($\pm$0.657)}\end{tabular}   & \textbf{\begin{tabular}[c]{@{}c@{}}0.814\\ \small{($\pm$0.060)}\end{tabular}}  \\
MTBench (Weather)   & \begin{tabular}[c]{@{}c@{}}3.920\\ \small{($\pm$0.084)}\end{tabular}   & \begin{tabular}[c]{@{}c@{}}3.531\\ \small{($\pm$0.186)}\end{tabular}   & \begin{tabular}[c]{@{}c@{}}4.091\\ \small{($\pm$0.245)}\end{tabular}   & \textbf{\begin{tabular}[c]{@{}c@{}}3.523\\ \small{($\pm$0.114)}\end{tabular}} & \begin{tabular}[c]{@{}c@{}}3.958\\ \small{($\pm$0.135)}\end{tabular}   & \begin{tabular}[c]{@{}c@{}}3.965\\ \small{($\pm$0.204)}\end{tabular}   & \begin{tabular}[c]{@{}c@{}}4.714\\ \small{($\pm$0.252)}\end{tabular}     & \begin{tabular}[c]{@{}c@{}}4.695\\ \small{($\pm$0.180)}\end{tabular}   & \begin{tabular}[c]{@{}c@{}}4.162\\ \small{($\pm$0.244)}\end{tabular}           \\
TSQA                & \begin{tabular}[c]{@{}c@{}}96.391\\ \small{($\pm$87.878)}\end{tabular} & \begin{tabular}[c]{@{}c@{}}47.524\\ \small{($\pm$26.683)}\end{tabular} & \begin{tabular}[c]{@{}c@{}}69.466\\ \small{($\pm$49.069)}\end{tabular} & \begin{tabular}[c]{@{}c@{}}51.368\\ \small{($\pm$34.393)}\end{tabular}        & \begin{tabular}[c]{@{}c@{}}81.097\\ \small{($\pm$40.683)}\end{tabular} & \begin{tabular}[c]{@{}c@{}}85.889\\ \small{($\pm$46.709)}\end{tabular} & \begin{tabular}[c]{@{}c@{}}453.377\\ \small{($\pm$121.495)}\end{tabular} & \begin{tabular}[c]{@{}c@{}}61.294\\ \small{($\pm$32.509)}\end{tabular} & \textbf{\begin{tabular}[c]{@{}c@{}}27.845\\ \small{($\pm$9.433)}\end{tabular}} \\

\emph{\textbf{Average}}             & \begin{tabular}[c]{@{}c@{}}33.774\\ \small{($\pm$29.396)}\end{tabular} & \begin{tabular}[c]{@{}c@{}}17.343\\ \small{($\pm$9.006)}\end{tabular}  & \begin{tabular}[c]{@{}c@{}}24.853\\ \small{($\pm$16.481)}\end{tabular} & \begin{tabular}[c]{@{}c@{}}19.229\\ \small{($\pm$12.061)}\end{tabular}        & \begin{tabular}[c]{@{}c@{}}28.711\\ \small{($\pm$13.637)}\end{tabular} & \begin{tabular}[c]{@{}c@{}}31.302\\ \small{($\pm$16.153)}\end{tabular} & \begin{tabular}[c]{@{}c@{}}164.795\\ \small{($\pm$42.323)}\end{tabular}  & \begin{tabular}[c]{@{}c@{}}22.521\\ \small{($\pm$11.115)}\end{tabular} & \textbf{\begin{tabular}[c]{@{}c@{}}10.940\\ \small{($\pm$3.246)}\end{tabular}} \\
\midrule
\multicolumn{10}{c}{\emph{Question Answering (Accuracy) - Higher is Better}} \\

MTBench (Finance)   & \begin{tabular}[c]{@{}c@{}}0.823\\ \small{($\pm$0.017)}\end{tabular}   & \begin{tabular}[c]{@{}c@{}}0.847\\ \small{($\pm$0.012)}\end{tabular}   & \begin{tabular}[c]{@{}c@{}}0.797\\ \small{($\pm$0.026)}\end{tabular}   & \begin{tabular}[c]{@{}c@{}}0.817\\ \small{($\pm$0.025)}\end{tabular}          & \begin{tabular}[c]{@{}c@{}}0.750\\ \small{($\pm$0.008)}\end{tabular}   & \begin{tabular}[c]{@{}c@{}}0.747\\ \small{($\pm$0.021)}\end{tabular}   & \begin{tabular}[c]{@{}c@{}}0.823\\ \small{($\pm$0.046)}\end{tabular}     & \begin{tabular}[c]{@{}c@{}}0.773\\ \small{($\pm$0.026)}\end{tabular}   & \textbf{\begin{tabular}[c]{@{}c@{}}0.913$^{\dagger}$\\ \small{($\pm$0.012)}\end{tabular}}  \\
MTBench (Weather)   & \begin{tabular}[c]{@{}c@{}}0.617\\ \small{($\pm$0.048)}\end{tabular}   & \begin{tabular}[c]{@{}c@{}}0.653\\ \small{($\pm$0.026)}\end{tabular}   & \begin{tabular}[c]{@{}c@{}}0.600\\ \small{($\pm$0.037)}\end{tabular}   & \begin{tabular}[c]{@{}c@{}}0.617\\ \small{($\pm$0.021)}\end{tabular}          & \begin{tabular}[c]{@{}c@{}}0.520\\ \small{($\pm$0.036)}\end{tabular}   & \begin{tabular}[c]{@{}c@{}}0.500\\ \small{($\pm$0.022)}\end{tabular}   & \begin{tabular}[c]{@{}c@{}}0.593\\ \small{($\pm$0.031)}\end{tabular}     & \begin{tabular}[c]{@{}c@{}}0.613\\ \small{($\pm$0.039)}\end{tabular}   & \textbf{\begin{tabular}[c]{@{}c@{}}0.717\\ \small{($\pm$0.052)}\end{tabular}}  \\
TSQA                & \begin{tabular}[c]{@{}c@{}}0.727\\ \small{($\pm$0.045)}\end{tabular}   & \begin{tabular}[c]{@{}c@{}}0.673\\ \small{($\pm$0.058)}\end{tabular}   & \begin{tabular}[c]{@{}c@{}}0.710\\ \small{($\pm$0.045)}\end{tabular}   & \begin{tabular}[c]{@{}c@{}}0.713\\ \small{($\pm$0.046)}\end{tabular}          & \begin{tabular}[c]{@{}c@{}}0.760\\ \small{($\pm$0.062)}\end{tabular}   & \begin{tabular}[c]{@{}c@{}}0.693\\ \small{($\pm$0.048)}\end{tabular}   & \begin{tabular}[c]{@{}c@{}}0.580\\ \small{($\pm$0.051)}\end{tabular}     & \begin{tabular}[c]{@{}c@{}}0.677\\ \small{($\pm$0.039)}\end{tabular}   & \textbf{\begin{tabular}[c]{@{}c@{}}0.870$^{\dagger}$\\ \small{($\pm$0.059)}\end{tabular}}  \\

\emph{\textbf{Average}}             & \begin{tabular}[c]{@{}c@{}}0.722\\ \small{($\pm$0.037)}\end{tabular}   & \begin{tabular}[c]{@{}c@{}}0.724\\ \small{($\pm$0.032)}\end{tabular}   & \begin{tabular}[c]{@{}c@{}}0.702\\ \small{($\pm$0.036)}\end{tabular}   & \begin{tabular}[c]{@{}c@{}}0.716\\ \small{($\pm$0.031)}\end{tabular}          & \begin{tabular}[c]{@{}c@{}}0.677\\ \small{($\pm$0.035)}\end{tabular}   & \begin{tabular}[c]{@{}c@{}}0.647\\ \small{($\pm$0.030)}\end{tabular}   & \begin{tabular}[c]{@{}c@{}}0.666\\ \small{($\pm$0.043)}\end{tabular}     & \begin{tabular}[c]{@{}c@{}}0.688\\ \small{($\pm$0.034)}\end{tabular}   & \textbf{\begin{tabular}[c]{@{}c@{}}0.833$^{\dagger}$\\ \small{($\pm$0.041)}\end{tabular}}  \\ 
\bottomrule
\end{tabular}%
}
\caption{Overall performance comparison on the three benchmarks. Best results are highlighted in \textbf{bold}. $^{\dagger}$ indicates that \ours{} significantly outperforms the strongest baseline\,(paired McNemar, $p<0.05$; \S\ref{app:significance}).} \label{table:main_results}
\vspace*{-0.4cm}
\end{table*}

\subsection{Setup} \label{section:exp_setup}

\paragraph{Benchmarks and Tasks} As presented in \autoref{table:benchmark}, we evaluate on three public benchmarks totaling \emph{20 tasks} that jointly cover context-grounded multimodal reasoning, sensor classification, and multi-task time-series QA. \textbf{MTBench}\,\citep{MTBench_25} is a multimodal benchmark coupling time series with rich textual context. It comprises 9 tasks across finance and weather, including trend classification, forecasting and indicator prediction, correlation classification, and multiple-choice QA. Inputs use either a short-window or long-window setting (e.g., different granularity regimes), which stresses temporal reasoning under varying horizons. \textbf{TimerBed}\,\citep{VLTime_NAACL_25} focuses on classification from diverse sensor domains; we evaluate on 6 datasets spanning 2-7 classes and sequence lengths up to 4K, emphasizing pattern recognition under domain shift. \textbf{TSQA}\,\citep{TimeMQA_ACL_25} is a multi-task benchmark spanning anomaly detection, classification, forecasting, imputation, and open-ended QA, providing a broad testbed for time-series reasoning.

\paragraph{Evaluation Metrics} For classification and MCQA tasks, we report accuracy (and weighted F1 when applicable). For regression tasks, we report MAE with MSE or MAPE as a secondary metric. For open-ended QA, we report an automatic NLI-based correctness score (entailment accuracy) computed with a DeBERTa-v3-base NLI classifier, used as a scalable proxy for human judgment. \S\ref{app:scorer_validation} validates this scorer against blinded human annotation and an independent LLM-judge panel. We report mean$\pm$std over 3 runs on cost-aware sampled test subsets (see \S\ref{section:data_desc}).

\paragraph{Comparison Baselines} We compare against single-agent prompting, multi-agent debate, and time-series-specific reasoning baselines. Single-agent baselines include \textbf{Zero-Shot} and Chain-of-Thought\,(\textbf{CoT})~\citep{CoT_NeurIPS_22}, each evaluated in both standard and multimodal settings (\textbf{+MM}). For debate-based methods, we include \textbf{MAD}~\citep{MAD_EMNLP_24} and its multimodal variant. For time-series reasoning baselines, we include \textbf{VL-Time}~\citep{VLTime_NAACL_25} and \textbf{ByMyEyes}~\citep{ByMyEyes_EMNLP_24}, which are designed specifically for time series. To ensure that gains are not simply artifacts of tool access, we additionally evaluate tool-augmented variants\,(\textbf{+Tools}), equipping the zero-shot, CoT, and MAD multimodal baselines with the same numerical lookup and code-execution interfaces available to \ours{}.

\paragraph{Implementation Details} We use \texttt{gpt-4.1-mini} as the backbone model via API, chosen for its strong zero-shot performance at favorable cost and latency (see \S\ref{section:backbone_prelims}). We use temperature $=0$ with fixed seeds. \ours{} uses a fixed configuration with $2$ analyst rounds, $3$ parallel reviewers, and a final synthesizer. For multimodal inputs, we provide two figures: a multi-panel time-domain plot and a multi-panel frequency-domain plot. All methods share the same backbone and evaluation protocol for fair comparison.

\subsection{Main Results} \label{section:main_results}

\autoref{table:main_results} summarizes performance on 20 tasks (mean$\pm$std over 3 runs), reported by benchmark and task type. Tasks are grouped into \textbf{classification}, \textbf{regression}, and \textbf{QA}: MTBench includes trend and correlation classification, forecasting and indicator prediction for regression, and MCQA; TimerBed averages six classification tasks; and TSQA combines classification and anomaly detection, forecasting and imputation (regression), and open-ended QA. We focus on the training-free regime, evaluating whether structured inference-time reasoning can extract more from off-the-shelf LLMs than current zero-shot approaches. Across all three task types, \ours{} delivers consistent gains over the strongest baseline, 25.65\% on classification, 36.92\% on regression error reduction, and 15.06\% on QA on average, without any task-specific training. Detailed experimental results are presented in \S\ref{section:extra_results}.

\vspace*{-0.1cm}
\paragraph{Classification} \ours{} achieves the highest average accuracy across all benchmarks, with a \textbf{25.65\%} improvement over the strongest baseline on average and up to \textbf{80.41\%} over standard zero-shot prompting. Improvements are consistent on MTBench (Finance/Weather), TimerBed, and TSQA, indicating effective temporal reasoning across diverse classification settings.

\vspace*{-0.1cm}
\paragraph{Regression} \ours{} obtains the lowest average MAE on MTBench (Finance) and TSQA, reducing error by \textbf{36.92\%} on average relative to the best baseline and avoiding catastrophic failures observed in naive multimodal variants. However, \ours{} does not dominate all regression regimes. On MTBench (Weather), CoT+MM performs slightly better (3.523 vs.\ 4.162 MAE), suggesting that some point-wise prediction settings benefit more from direct multimodal prompting than multi-agent deliberation.

\paragraph{Question Answering} \ours{} achieves the highest average QA accuracy across benchmarks, with \textbf{15.06\%--28.75\%} improvements over prior methods. These gains suggest more effective integration of temporal evidence from multiple aspects of time series for context-grounded QA.

 \begin{figure*}[t]
    \centering
    \includegraphics[width=\linewidth]{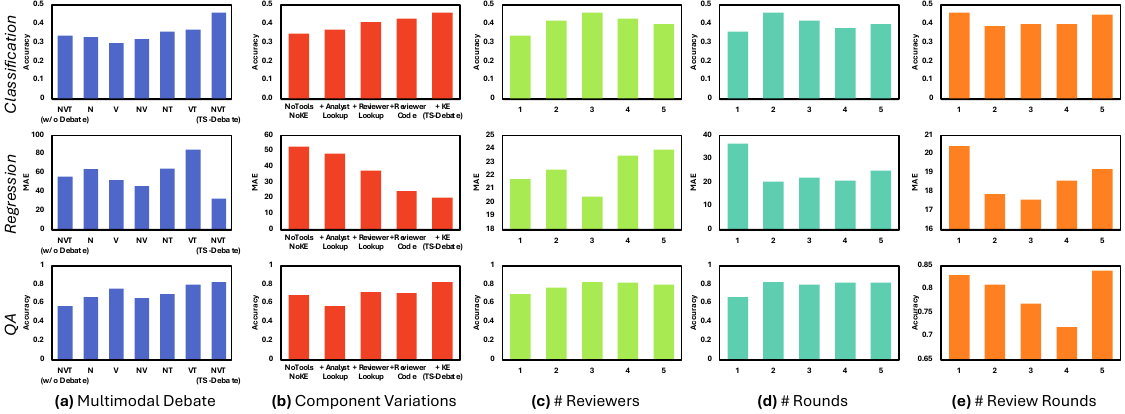}
    \vspace*{-0.5cm}
    \caption{Impact of (a) modality composition, (b) system components, and (c--e) debate configurations across tasks.} \label{figure:ablation_results}
    \vspace*{-0.2cm}
\end{figure*}

\paragraph{Statistical Significance} We test each benchmark $\times$ task-type cell of \autoref{table:main_results} against its strongest baseline using instance-level paired tests\,(exact McNemar for accuracy; sign tests for regression). \ours{}'s pooled gains are significant for classification\,($p{=}3.6\times10^{-24}$; $n{=}2{,}931$) and QA\,($p{=}3.4\times10^{-8}$; $n{=}731$), while regression differences are task-dependent and not significant in aggregate. \ours{} has the better point estimate on 18 of 20 tasks, with six significant improvements and significant declines only on the two local-value-reconstruction tasks\,(weather point forecasting and TSQA imputation; \S\ref{app:error_analysis}). Task-level tests and robustness checks, including cross-run deduplication, run-level replication, and comparisons with all eight baselines, appear in \S\ref{app:significance}.

These results show that \ours{} is strongest when the target depends on global structure, cross-view consistency, or relative quantities, rather than purely local reconstruction.

\subsection{Ablation and Sensitivity Analysis} \label{section:ablation_results}
We conduct controlled ablations on 20 instances per task (400 samples total) to isolate the effects of modality composition, system components, and debate configurations.

\paragraph{Multimodal Debate Agents} \autoref{figure:ablation_results}(a) shows that incorporating multiple modalities consistently improves performance over single-modality or no-debate\,(i.e., simple agents with tools) baselines. While numerical (\textsf{N}), visual (\textsf{V}), and textual (\textsf{T}) analysts each contribute complementary gains, their combination (\textsf{NVT}) yields the best results across all tasks. Notably, removing debate leads to substantial degradation, confirming that performance gains stem from structured interaction rather than modality access alone.

\paragraph{Component Variations} \autoref{figure:ablation_results}(b) demonstrates improvements as key components are added. Analyst-side lookup reduces error and improves accuracy, reviewer-side lookup and code execution further refine predictions, and knowledge elicitation (\textsf{KE}) provides a substantial marginal gain, yielding the strongest overall performance. These results show that both external tool use and structured reasoning are essential to effective debate.

\paragraph{Debate Configurations} \autoref{figure:ablation_results}(c--e) analyze sensitivity to the number of reviewers, debate rounds, and review rounds. Performance improves with additional reviewers and rounds up to a moderate regime (typically 2--3), after which gains saturate or slightly regress, especially for regression tasks. This indicates diminishing returns and supports our default configuration as a favorable trade-off between performance and efficiency.

As VCC is not a separable module but is embedded in the reviewer/synthesizer protocol and output structure\,(e.g., verification tags, conflict status, and calibration rules), directly removing it would require redesigning prompts and decision logic, resulting in a qualitatively different method and confounding attribution. Instead, \emph{we examine VCC mechanisms via tool and modality ablations}. Concretely, no tools in \autoref{figure:ablation_results}(b) disables programmatic verification, and single-modality settings in \autoref{figure:ablation_results}(a) limit cross-modal conflict resolution, thereby probing the key VCC mechanisms.

\begin{table}[t]
\centering
\resizebox{\linewidth}{!}{%
\begin{tabular}{@{}l|ccc|cc@{}}
\toprule
\textbf{Method}    & \textbf{Regression $\downarrow$} & \textbf{Classification  $\uparrow$} & \textbf{QA  $\uparrow$} & \textbf{\# Tokens} & \textbf{Time (s)} \\ \midrule \midrule
ZS+MM+Tools  & 23.32 & 0.29 & 0.58 & 32955 & 31.9 \\
CoT+MM+Tools & 25.98 & 0.33 & 0.63 & 30899 & 28.7 \\
MAD+MM+Tools & 24.77 & 0.39 & 0.67 & 82488 & 63.9 \\ \midrule
\emph{\ours{}} & \textbf{20.46}        & \textbf{0.47}             & \textbf{0.83} & 77002             & 77.3                   \\ \bottomrule
\end{tabular}%
}
\caption{Result comparison on the tool-matched setup.} \label{table:tool_match}
\vspace*{-0.35cm}
\end{table}

\paragraph{Tool-Matched Setups} To test whether \ours{}'s gains are explained by tool access rather than by the protocol itself, \autoref{table:tool_match} reports results on the ablation set where ZS+MM, CoT+MM, and MAD+MM baselines all receive the same lookup and code-execution interface as \ours{}. Tool access alone does not recover \ours{}: even with identical tools and comparable or larger token budgets\,(MAD+MM+Tools uses 82.5K vs.\ 77.0K tokens), prior methods trail \ours{} by $+0.08$ to $+0.18$ on classification and $+0.16$ to $+0.25$ on QA. This indicates that the gains arise from modality-specialized evidence generation and the VCC protocol, not from unstructured tool calling.

\paragraph{Conflict Resolution Analysis} Across all 5{,}742 stored traces, reviewer agents issued 28{,}741 claim verdicts: 94.5\% verified, 3.0\% unverified, 1.6\% contradicted, and 0.9\% partially verified. On discrete tasks, conflict-flagged samples are \emph{more} accurate than conflict-free samples (0.625, $n{=}547$ vs.\ 0.536, $n{=}3{,}095$), suggesting that explicit verification resolves cross-modal disagreement rather than propagating it. Together with the matched setups above, these results attribute the gains to verification-driven conflict resolution rather than ensembling. \S\ref{app:case_studies} presents two verbatim traces illustrating a success and a failure.

\subsection{Resource Cost} \label{section:resource_cost}

\begin{figure}[H]
    \centering
    \includegraphics[width=\linewidth]{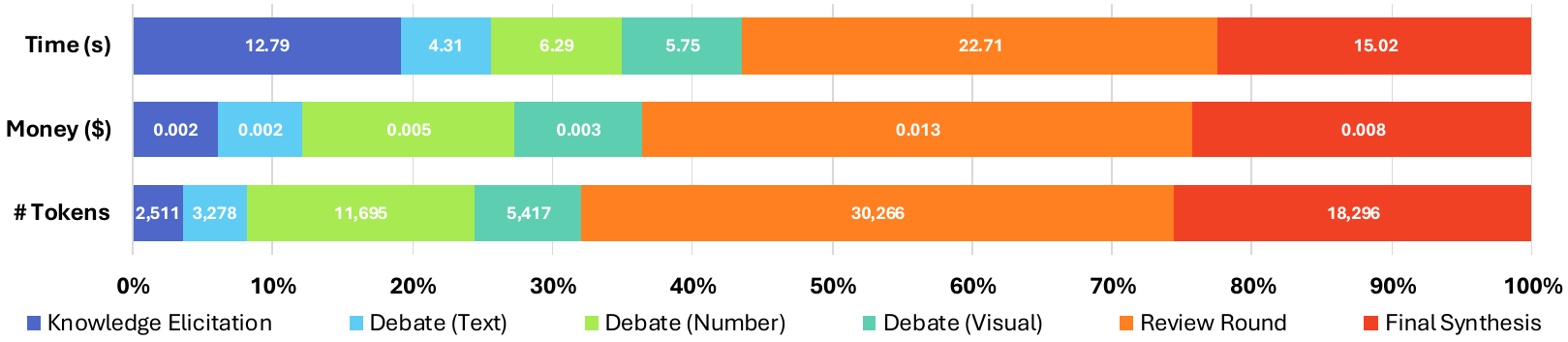}
    \caption{Average time and monetary cost breakdown.} \label{figure:cost_breakdown}
    \vspace*{-0.3cm}
\end{figure}

\autoref{figure:cost_breakdown} and \autoref{table:cost_comparison} report wall-clock time, token usage, and monetary API cost, averaged across tasks for a single run using closed-source LLMs. Under the default configuration (\texttt{gpt-4.1-mini}), \ours{} requires about 71 seconds and incurs a cost of \$0.033 per sample. The overhead from time-frequency preprocessing and chart generation is negligible and can be handled offline. Agents equipped with tools tend to consume more tokens, likely due to iterative tool invocations; however, preliminary trials suggest that stronger models can reduce overall token usage by making more accurate and efficient tool calls, thereby converging with fewer errors and interactions.

Cost is controllable along three axes. Reducing reviewer agents from three to one cuts token usage by ${\approx}45\%$ with only a moderate accuracy decrease (\autoref{figure:ablation_results}(c)); debate rounds beyond two add 13--35\% more tokens without consistent gains (\autoref{figure:ablation_results}(d)), making two rounds the efficiency knee; and dropping the visual channel yields a text$+$numerical configuration that runs on any text-only LLM while retaining most of the benefit (classification 0.373 vs.\ 0.408; QA 0.702 vs.\ 0.719). The conflict signal further suggests a conflict-aware dynamic stopping rule---ending debate early when reviewers detect no conflict and allocating additional review only when a conflict is flagged---which we leave to future work. \ours{} is thus intended for error-sensitive, auditable offline analysis rather than high-throughput serving.
\vspace*{-0.1cm}
\section{Conclusions} \label{section:conclusions}
\vspace*{-0.1cm}
We presented \ours{}, an inference-time protocol for \emph{zero-shot} TSR. Modality-specialized text, visual, and numerical analysts produce structured evidence, and a VCC protocol verifies temporally checkable claims via code and lookup, types cross-modal disagreement, and calibrates answers by temporal scope. Across 20 tasks on three benchmarks, \ours{} outperforms strong single-agent, multi-agent, and tool-matched baselines, with the largest gains where reasoning depends on global structure, cross-view consistency, or numerical faithfulness rather than local reconstruction. The bottleneck for zero-shot TSR is not modality access but the \emph{organization} of evidence, verification, and calibration at inference time.


\section*{Limitations}
\ours{} has several limitations that suggest directions for future work. Performance can degrade for long-horizon causal reasoning or highly noisy multivariate time series where visual trends and numerical signals diverge. Domain knowledge elicitation is currently sample-specific, and numeric verification is constrained by the expressiveness of available tools. Promising directions include transferable and adaptive knowledge elicitation across related tasks, debate topologies that respond to disagreement patterns, richer verification\,(e.g., robust statistical tests), and tighter integration with structured retrieval for contextual grounding.

Multi-agent orchestration also increases token usage and wall-clock time relative to single-pass prompting. \ours{} provides cost controls via modality selection, the number of reviewers, and the number of debate rounds, but aggressive pruning reduces verification coverage; adaptive topologies that allocate reviewers only when conflicts are detected are another promising direction. Accordingly, \ours{} is not intended as a universal replacement for lightweight classification heuristics, but rather for high-stakes, offline, or error-sensitive deployments where fine-tuning is infeasible due to data scarcity and where transparent, explicitly verified reasoning justifies the computational overhead.

Our experiments deliberately fix a single fast, non-reasoning backbone across all methods so that gains are attributable to the inference-time protocol rather than to model-specific internal reasoning (\S\ref{section:backbone_prelims}); whether the benefits transfer to reasoning-centric or substantially stronger backbones remains open, as does assigning heterogeneous backbones to different analysts, which may make disagreement itself more informative. Our evaluation covers three public benchmarks; complementary suites such as TSR-Suite~\citep{TimeOmni1}, which targets multi-level temporal reasoning in a text+numerical setting, are a natural extension. Finally, TS-Debate is training-free by design: recent training-based systems such as TimeART~\citep{TimeART_2026} and TimeOmni-1~\citep{TimeOmni1} pursue related goals through fine-tuning and reinforcement learning, and comparing against them where training-free variants exist is left to future work.

\section*{Acknowledgments}
This work was conducted as part of the research activities at DeepAuto.ai. This work was supported by Institute for Information \& communications Technology Planning \& Evaluation\,(IITP) grant funded by the Korea government\,(MSIT) (RS-2019-II190075, Artificial Intelligence Graduate School Program\,(KAIST)), and National Research Foundation of Korea\,(NRF) grant funded by the Korea government\,(MSIT) (No. RS-2023-00256259).


\bibliography{6_References}

\newpage
\appendix
\section{Extended Related Work} \label{section:more_related_work}

\subsection{Time-Series Analysis and Reasoning}
Recent work explores using LLMs for time-series analysis and reasoning. A direct approach serializes numerical sequences as text, enabling zero-shot forecasting but suffering from sensitivity to tokenization and brittle numerical fidelity\,\citep{LLMTime_NeurIPS_23}. Alignment-based methods\,\citep{TimeLLM_ICLR_24,toward_tsr_NeurIPSW_24} address this issue by training adapters or auxiliary encoders that map time series into LLM-compatible representations, improving performance at the cost of modality-specific fine-tuning and additional training data. Several systems further couple LLMs with specialized time-series models. TS-Reasoner\,\citep{TSReasoner_25} integrates foundation time-series encoders with LLMs to support multi-step workflows over forecasting, anomaly detection, and causal discovery. ChatTime\,\citep{ChatTime_AAAI_25} treats time series as a ``foreign language,'' reducing tokenization overhead while supporting bi-modal input and output. Despite their effectiveness, these approaches primarily focus on representation learning or encoder design and typically operate within a single numerical information channel. Recent training-based systems push TSR further. TimeART~\citep{TimeART_2026} trains agentic, tool-augmented time-series reasoning, and TimeOmni-1~\citep{TimeOmni1} instills perception, extrapolation, and decision-making capabilities through supervised fine-tuning and task-grounded reinforcement learning, introducing the TSR-Suite benchmark. Both require task-specific training and therefore sit outside our training-free, inference-time scope.

In contrast, \ours{} assumes off-the-shelf LLM and MLLM backbones and targets an \emph{inference-time orchestration protocol}. Rather than training new encoders, we coordinate multiple modality-specialized agents (numerical, visual, and textual) under explicit verification and calibration rules. This design avoids task-specific training and instead emphasizes reliability and numerical faithfulness through structured, branch-style reasoning at inference time. \ours{} is complementary in that its verification-driven protocol needs no parameter updates and could, in principle, orchestrate such fine-tuned models as analysts.

\subsection{Multimodal LLMs for Time Series} 
A parallel line of work leverages multimodal LLMs\,(MLLMs) by converting time series into richer visual or cross-modal representations. VL-Time\,\citep{VLTime_NAACL_25} proposes a visualization-first pipeline that transforms time series into images\,(e.g., waveforms and spectrograms) and reasons over them with MLLMs, achieving substantial gains with reduced token budgets. ByMyEyes\,\citep{ByMyEyes_EMNLP_24} similarly shows that visual prompting over sensor-derived plots can outperform text-only prompts while further lowering token costs. While these studies demonstrate the effectiveness of multimodal representations, they typically rely on a \emph{single fusion step} within an MLLM\,\citep{EmpoweringTSR_MLLM_25}. \ours{} departs from this paradigm in two key ways. First, instead of learning new multimodal encoders, we keep the underlying models fixed and instantiate modality-isolated agents for numerical series, visual patterns, and textual context. Second, we coordinate these agents through a multi-stage collaborative protocol with explicit verification, conflict resolution, and calibration, allowing cross-modal evidence to interact iteratively rather than being fused in a single forward pass.

\subsection{Multi-Agent Debate} 
Multi-agent debate\,(MAD) frameworks\,\citep{MAD_Strategy_ICML_24} employ multiple LLM agents that iteratively critique one another, often moderated by a judge, to improve reasoning quality. Early work demonstrated gains on mathematical and logical benchmarks, while later studies examined debate topology and aggregation rules and cautioned against over-interpreting debate-based evaluations\,\citep{SparseMAD_EMNLP_24,Vote_vs_Consensus_ACL_25,StopOvervaluing_25}. These approaches are predominantly text-only and typically assume homogeneous agents with shared access to the same information. \ours{} differs from generic MAD in both \emph{assumptions} and \emph{architecture}. Rather than adversarial persuasion or majority voting, \ours{} adopts a \emph{collaborative} debate paradigm in which agents contribute complementary evidence. We enforce \emph{modality isolation} by assigning each agent a distinct view of the problem, ensuring that disagreements reflect genuinely different evidence rather than paraphrased reasoning. Moreover, our judges are equipped with explicit verification tools, including code execution, to check numerical claims and domain constraints. As a result, debate in \ours{} is not an evaluation heuristic but an inference-time protocol tailored to multimodal TSR, explicitly targeting numerical hallucinations and calibration errors. \autoref{table:framework_comparison} summarizes the key differences between \ours{} and existing frameworks.

\section{Pseudocode of \ours{}} \label{section:pseudocode}
We present the pseudocode of the proposed \ours{} framework in Algorithm~\ref{alg:pseudo_code} below. The source code is available at \url{https://github.com/DeepAuto-AI/TS-Debate}.
\begin{algorithm}[H]
\caption{Overall Procedure of \ours{}} \label{alg:pseudo_code}
\begin{algorithmic}[1]
\REQUIRE Task query $q$, time series $\mathbf{x}_{1:T}$, textual context $c$, number of debate rounds $R$, number of reviewer agents $J$
\ENSURE Final calibrated answer $\hat{a}$

\COMMENT{Multimodal Query Construction (\S\ref{section:mqc})}
\STATE $k \gets \mathsf{Elicit}(q,c)$
\STATE $\mathsf{Chart}_{\textrm{time}} \gets \mathsf{ChartGen}_{\textrm{time}}(\mathbf{x}_{1:T})$;\;\;
       $\mathsf{Chart}_{\textrm{freq}} \gets \mathsf{ChartGen}_{\textrm{freq}}(\mathbf{x}_{1:T})$
\STATE $I_{\textsf{V}} \gets [\mathsf{Chart}_{\textrm{time}}, \mathsf{Chart}_{\textrm{freq}}]$;\;\;
       $I_{\textsf{N}} \gets \mathsf{Lookup}(\mathbf{x}_{1:T})$;\;\;
       $I_{\textsf{T}} \gets c$

\COMMENT{Multimodal Collaborative Debate (\S\ref{section:main_debate})}
\STATE $E^{(0)} \gets \varnothing$
\FOR{$r=1$ to $R$}
    \FORALL{$m \in \mathcal{M}$}
        \STATE $e_m^{(r)} \gets \mathcal{A}_m\!\left(q, I_m, k, E^{(r-1)}\right)$
    \ENDFOR
    \STATE $E^{(r)} \gets \{e_m^{(r)}\}_{m\in\mathcal{M}}$
\ENDFOR

\COMMENT{Verification-Conflict-Calibration (VCC) Protocol (\S\ref{section:vcc_protocol})}
\STATE $E^{(R)} \gets \{e_m^{(R)}\}_{m\in\mathcal{M}}$
\STATE $I \gets [I_{\textsf{T}}, I_{\textsf{V}}, I_{\textsf{N}}]$
\FOR{$j=1$ to $J$}
    \STATE $u_j \gets \mathsf{VCC}_j(q, I, k, E^{(R)})$
\ENDFOR
\STATE $\hat{a} \gets \mathsf{VCC}_{\textrm{final}}(q, I, k, \{u_j\}_{j=1}^J)$
\STATE \textbf{return} $\hat{a}$
\end{algorithmic}
\end{algorithm}

\section{Implementation Details} \label{section:implementation_details}

\subsection{Dataset Descriptions} \label{section:data_desc}

We evaluate our approach on three public benchmarks comprising a total of 20 tasks. These benchmarks are selected to stress complementary aspects of time-series reasoning, including multimodal grounding with rich textual context, sensor-based classification under domain shift, and broad multi-task question answering\,(QA) over time-series data.

\begin{itemize}[leftmargin=9pt, nosep, noitemsep]
    \item \textbf{MTBench}~\citep{MTBench_25}. MTBench is a multimodal time-series benchmark that pairs numeric sequences with rich textual context drawn from financial news and weather reports. The dataset is publicly available at \url{https://github.com/Graph-and-Geometric-Learning/MTBench}. We evaluate nine tasks spanning finance and weather domains, including trend classification, forecasting, indicator prediction, correlation classification, and multiple-choice question answering\,(MCQA). MTBench includes both short- and long-window settings with distinct temporal granularity regimes, e.g., 1-hour resolution with 7-day input windows and 5-minute resolution with 30-day input windows.

    \item \textbf{TimerBed}~\citep{VLTime_NAACL_25}. TimerBed is a diverse time-series classification benchmark covering multiple sensor-based domains, including healthcare, human activity recognition\,(HAR), bioacoustics, energy, and geophysics. The benchmark is available at \url{https://github.com/AdityaLab/TimerBed}. We evaluate six datasets with varying sequence lengths and between 2 and 7 classes. Specifically, these include ECG signals for cardiac rhythm classification (4 classes), HAR using 3-axis accelerometer data (6 classes), EMG signals for muscle activity classification (3 classes), CTU energy usage data (2 classes), RCW whale call detection (2 classes), and TEE lightning event classification (7 classes).

    \item \textbf{TSQA}~\citep{TimeMQA_ACL_25}. TSQA is a large-scale, multi-task benchmark for time-series question answering, available at \url{https://huggingface.co/datasets/Time-MQA/TSQA}. It spans a broad range of task types, including anomaly detection, classification, forecasting, imputation, and open-ended QA. The benchmark supports multiple output formats, such as open-ended generation, multiple-choice questions, and true/false queries. TSQA covers diverse application domains, with major categories including Health/Healthcare, HAR, Web data, Nature, and Environment/AIOps.
\end{itemize}

\subsection{Sampling Strategy}
Running multi-agent LLM methods on the full evaluation splits (over 200K instances across all tasks) is cost-prohibitive. We therefore use a \emph{cost-aware} but \emph{distribution-preserving} sampling protocol that caps per-task evaluation instances and stratifies sampling along the benchmark-defined axes most correlated with task difficulty.

\paragraph{Main Experiments} For each task and each run, we sample $n = \min(100, \; \texttt{available\_test\_instances})$ instances. Tasks with fewer than 100 instances (e.g., EMG: 41, TEE: 73) use the full available set.

\paragraph{Ablation and Hyperparameter Studies} For each task, we sample $20$ instances and run each ablation configuration once, using a fixed seed.

\paragraph{Stratification} We apply benchmark-specific stratification as follows.
\begin{itemize}[leftmargin=9pt, nosep, noitemsep]
    \item \textbf{MTBench (trend/forecasting/indicator/correlation):} stratify by input source (short-window 7d vs.\ long-window 30d).
    \item \textbf{MTBench (MCQA):} random sampling (uniform).
    \item \textbf{TimerBed:} hybrid stratification by class label (class-balanced) and sequence length buckets (difficulty proxy).
    \item \textbf{TSQA (anomaly/classification):} stratify by domain.
    \item \textbf{TSQA (forecasting/imputation):} difficulty-stratify by input length.
    \item \textbf{TSQA (QA):} stratify by domain and question type (open-ended, multiple choice, and True/False).
\end{itemize}

\paragraph{Reproducibility} We set temperature $=0$ and seed both sample selection and LLM calls. Sample selection uses \texttt{seed} $=2025 + \texttt{run\_id}$ (e.g., run 1 uses seed 2026).

\paragraph{Sample Representativeness} We validate sampling effects using chi-squared ($\chi^2$) tests where applicable. For TSQA, stratified sampling preserves the original task distribution ($\chi^2=1.38$, $p=0.85$). For MTBench, capping at 100 samples per task intentionally balances task representation across finance and weather domains, while preserving original proportions. For TimerBed, we employ hybrid class-balanced sampling with difficulty stratification to ensure representative coverage of all class labels within each dataset.

\begin{table*}[t]
\centering
\resizebox{\textwidth}{!}{%
\begin{tabular}{@{}lccccccccc@{}}
\toprule
\textbf{Task} & \textbf{Domain} & \textbf{\begin{tabular}[c]{@{}c@{}}\# Samples\\ (Train/Test)\end{tabular}} & \textbf{Input Length} & \textbf{Output Length ($H$)} & \textbf{Dim} & \textbf{Granularity} & \textbf{Task Type} & \textbf{Labels/Classes} & \textbf{Metrics} \\ 
\midrule
\midrule
\multicolumn{10}{c}{\emph{MTBench (MIT License, \citet{MTBench_25})}} \\
\midrule
Trend Prediction & \multirow{5}{*}{Finance} & 1,275 & 115-755 & N/A & \multirow{9}{*}{1} & 1 hour / 5 min & Classification & \textless{}-4\%, -2\% $\sim$-4\%, -2\% $\sim$+2\%, +2\% $\sim$+4\%, \textgreater{}+4\% & Accuracy, F1 \\
Price Forecasting &  & 1,275 & 115-755 & 1-289 &  & 1 hour / 5 min & Regression & Continuous (next $H$ values) & MAE, MAPE \\
Technical Indicator &  & 1,275 & 115-755 & 1-289 &  & 1 hour / 5 min & Regression & Continuous (next $H$ values) & MAE, MAPE \\
Multiple Choice Question Answering &  & 1,007 & 114-402 & N/A &  & \multirow{6}{*}{1 hour} & QA & A, B, C, D & Accuracy, F1 \\
Correlation &  & 1,007 & 114-402 & N/A &  &  & Classification & \begin{tabular}[c]{@{}c@{}}Strong Positive, Moderate Positive, No Correlation, \\ Moderate Negative, Strong Negative\end{tabular} & Accuracy, F1 \\
Trend Prediction & \multirow{4}{*}{Weather} & 3,918 & 168-336 & N/A &  &  & Classification & increasing, decreasing, stable & Accuracy, F1 \\
Temperature Forecasting &  & 3,918 & 168-336 & 24-72 &  &  & Regression & Continuous (next H values) & MAE, MSE \\
Technical Indicator &  & 3,918 & 168-336 & 3 &  &  & Regression & max, min, diff (3 values) & MAE, MSE \\
Multiple Choice Question Answering &  & 1,373 & 168-336 & N/A &  &  & QA & A, B, C, D & Accuracy, F1 \\
\midrule
\multicolumn{10}{c}{\emph{TimerBed (\citet{VLTime_NAACL_25})}} \\
\midrule
CTU & Energy Usage & \begin{tabular}[c]{@{}c@{}}500 \\ (250/250)\end{tabular} & 720 & \multirow{6}{*}{N/A} & 1 & 2 min & \multirow{6}{*}{Classification} & desktop, laptop & \multirow{6}{*}{Accuracy, F1} \\
ECG & Healthcare & \begin{tabular}[c]{@{}c@{}}50,536 \\ (49,960/576)\end{tabular} & 1,500 &  & 1 & 100 Hz &  & \begin{tabular}[c]{@{}c@{}}normal sinus rhythm, fibrillation, \\ alternative rhythm, too noisy to be classified\end{tabular} &  \\
EMG & Healthcare & \begin{tabular}[c]{@{}c@{}}204 \\ (163/41)\end{tabular} & 1,500 &  & 1 & 4 kHz &  & healthy, suffering from neuropathy, suffering from myopathy &  \\
HAR & Human Activity & \begin{tabular}[c]{@{}c@{}}10,299 \\ (7,352/2,947)\end{tabular} & 206 &  & 3 & 50 Hz &  & \begin{tabular}[c]{@{}c@{}}walking, walking upstairs, walking downstairs, \\ sitting, standing, laying down\end{tabular} &  \\
RCW & Bioacoustics & \begin{tabular}[c]{@{}c@{}}12,896 \\ (10,934/1,962)\end{tabular} & 4,000 &  & 1 & 2 kHz &  & no right whale call, right whale call present &  \\
TEE & Geophysics & \begin{tabular}[c]{@{}c@{}}143 \\ (70/73)\end{tabular} & 319 &  & 1 & 50 MHz &  & \begin{tabular}[c]{@{}c@{}}CG Positive, IR Negative, SR Subsequent, I Impulsive, \\ I2 Impulsive Pair, KM Gradual, O Off-record\end{tabular} &  \\
\midrule
\multicolumn{10}{c}{\emph{TSQA (Apache License 2.0, \citet{TimeMQA_ACL_25})}} \\
\midrule
Anomaly Detection & Multi-Domain & 37,000 & 8-64 & N/A & 1 & \multirow{5}{*}{Variable} & Classification & Normal Point, Anomaly Point & Accuracy \\
Classification & Human Activity & 37,000 & 9-30 & N/A & 1 &  & Classification & \begin{tabular}[c]{@{}c@{}}Jogging, Walking, Upstairs, Downstairs, \\ Sitting, Standing, Freeze, No freeze\end{tabular} & Accuracy \\
Forecasting & Multi-Domain & 37,390 & 65-248 & 8-32 & 1 &  & Regression & Continuous (next $H$ values) & MAE, MAPE \\
Imputation & Multi-Domain & 38,657 & 97-256 & 4-12 & 1 &  & Regression & Continuous (missing $H$ values) & MAE, MAPE \\
Question Answering & Multi-Domain & 36,829 & 8-60 & N/A & 1 &  & QA & Open-ended (52\%), Multiple choice (29\%), True/False (19\%) & Accuracy \\ 
\bottomrule
\end{tabular}%
}
\caption{Summary of downstream tasks and benchmark dataset statistics.} \label{table:benchmark}
\vspace*{-0.2cm}
\end{table*}

\subsection{Baselines}
We implement and evaluate eight baselines covering single-agent prompting, multi-agent deliberation, and time series-specific visual grounding. All baselines use the same backbone LLM and an identical chart generation pipeline for multimodal variants. This design ensures that performance differences arise from the underlying \emph{reasoning protocols}, rather than from confounding architectural or implementation factors.
\begin{itemize}[leftmargin=9pt, nosep, noitemsep]
    \item \textbf{Zero-Shot}. A single LLM invocation that directly produces the final answer. We report both time series-only and multimodal variants (\textbf{+MM}), where the latter gets the same textual context as well as time- and frequency-domain visualizations as other multimodal methods.
    
    \item \textbf{CoT} \citep{CoT_NeurIPS_22}. A single-agent chain-of-thought prompting baseline, evaluated in both time series-only and multimodal settings (\textbf{+MM)}. The prompt is intentionally kept minimal to avoid task-specific engineering and to preserve comparability with the zero-shot setting.
    
    \item \textbf{Multi-Agent Debate (MAD)} \citep{MAD_EMNLP_24}. A multi-agent deliberation framework in which multiple agents iteratively critique and refine each other's responses. We follow the standard debate protocol and introduce a multimodal extension (\textbf{+MM}) that supplies the same textual context as well as time- and frequency-domain charts, ensuring parity with the information available to \ours{}.
    
    \item \textbf{ByMyEyes} \citep{ByMyEyes_EMNLP_24}. A visual prompting framework for sensor data that employs a two-phase dynamic visualization selection: (1)~a planning phase that filters candidate visualization tools based on the task description, and (2)~a selection phase that identifies the optimal domain-specific visualization\,(e.g., ECG peak plots).
    
    \item \textbf{VL-Time} \citep{VLTime_NAACL_25}. A two-stage visualization-language approach for time-series classification: (1)~a \emph{planning stage} where the LLM selects between time-domain and frequency-domain visualization and generates reasoning hints, and (2)~a \emph{solving stage} that conditions on the selected chart and textual hints. 
\end{itemize}

\subsection{Framework Configurations}

\paragraph{Backbone LLM} Unless otherwise noted, all experiments use \texttt{gpt-4.1-mini} with temperature $=0$ as discussed in \S\ref{section:backbone_prelims}.

\paragraph{Modalities and Tools} \ours{} instantiates three modality-specialized analysts: \textbf{Text} (context grounding), \textbf{Visual} (pattern recognition from plots), and \textbf{Numerical} (precise value retrieval via lookup tools). Reviewers and the final synthesizer implement the VCC protocol and may call a Python code executor to check arithmetic and derived quantities.

\paragraph{Chart Generation} For each instance, we generate a task-aware time-series plot and (when applicable) a frequency-spectrum plot. Visual prompts are \emph{identical} across all multimodal methods.

\paragraph{Tooling Constraints} To control cost and prevent unrestricted computation, we cap the number of tool calls per agent type (small constants) and expose only read-only lookup functions (e.g., value queries, feature queries, frequency features) plus a restricted Python executor for verification.

\begin{table}[t]
\centering
\resizebox{\linewidth}{!}{%
\begin{tabular}{@{}lcccc@{}}
\toprule
\textbf{Scorer} & \textbf{\begin{tabular}[c]{@{}c@{}}Marks correct\\(1{,}521 outputs)\end{tabular}} & \textbf{\begin{tabular}[c]{@{}c@{}}Marks correct\\(gold-88)\end{tabular}} & \textbf{\begin{tabular}[c]{@{}c@{}}Agreement\\w/ human\end{tabular}} & \textbf{$\kappa$ w/ human} \\ \midrule
NLI (paper metric) & 81.1\% & 84.1\% & 0.818 & 0.52 \\
gemini-2.5-flash   & 40.3\% & 56.8\% & 0.841 & 0.67 \\
grok-4.3           & 43.3\% & 54.5\% & 0.795 & 0.58 \\
gemini-3-flash     & 59.5\% & 70.5\% & 0.909 & 0.79 \\
deepseek-v3.2      & 42.1\% & 50.0\% & 0.818 & 0.64 \\ \midrule
Human gold         & ---    & 68.2\% & ---   & --- \\ \bottomrule
\end{tabular}%
}
\caption{Scorer strictness and agreement with the human gold standard
($n{=}88$ items on which both annotators agree). NLI error direction vs.\
gold: recall 0.98, precision 0.80 (TP 59, FP 15, FN 1, TN 13).} \label{tab:scorer_strictness}
\end{table}

\section{Open-Ended QA Scorer Validation} \label{app:scorer_validation}
The NLI-based correctness score affects only the open-ended slice of one of our 20 tasks (52\% of TSQA QA). We validate it in three ways.

\paragraph{Exact-Match Control} Restricting TSQA QA to its exact-match formats removes the NLI scorer entirely; TS-Debate's gain persists (0.832 vs.\ 0.702 for the strongest baseline; paired McNemar, $p{=}3.3\times10^{-3}$, $n{=}131$).

\paragraph{Blinded Human Audit} Two annotators independently labeled 108 stratified items (12 per method across nine methods, drawn across all three runs), seeing exactly the prediction--reference pairs the scorer receives. Inter-annotator raw agreement is 0.815 ($\kappa{=}0.60$); the gold standard is the 88-item agreement subset. NLI--human agreement is 0.818 (bootstrap 95\% CI $[0.74, 0.90]$; $\kappa{=}0.52$), and the scorer's errors are one-sided lenient: recall 0.983, precision 0.797 (15 false positives, 1 false negative). All 15 lenient errors fall on baseline outputs, and the single strict error penalizes TS-Debate. Eight of the 15 false positives are numeric-calculation questions (e.g., mean, median, next value), where textual entailment cannot check arithmetic.

\paragraph{LLM-Judge Triangulation} Four LLM judges from three model families (all different from the backbone), each given only the reference answer and the candidate answer, re-scored all 1{,}521 stored open-ended outputs (Tables~\ref{tab:scorer_strictness} and~\ref{tab:scorer_permethod}). Judges agree with one another ($\kappa$ 0.61--0.77) and with the human gold ($\kappa$ 0.58--0.79) more than any of them agrees with NLI ($\kappa \le 0.36$), so the NLI proxy is lenient in absolute terms. Crucially, this leniency does not favor our method: TS-Debate ranks first of nine methods under three judges and second under the fourth and under NLI, while the NLI-preferred baseline (CoT+MM) is mid-rank under every judge. Relative standings are scorer-robust; absolute rates are scorer-dependent.

\begin{table}[t]
\centering
\resizebox{\linewidth}{!}{%
\begin{tabular}{@{}lccccc@{}}
\toprule
\textbf{Method} & \textbf{NLI} & \textbf{gemini-2.5} & \textbf{grok-4.3} & \textbf{gemini-3} & \textbf{deepseek-v3.2} \\ \midrule \midrule
Zero-Shot & 0.876 & 0.503 & 0.467 & 0.651 & 0.485 \\
ZS+MM     & 0.751 & 0.367 & 0.473 & 0.615 & 0.432 \\
CoT       & 0.858 & \textbf{0.544} & 0.521 & 0.692 & 0.503 \\
CoT+MM    & \textbf{0.911} & 0.402 & 0.467 & 0.609 & 0.426 \\
MAD       & 0.805 & 0.396 & 0.349 & 0.592 & 0.402 \\
MAD+MM    & 0.751 & 0.284 & 0.320 & 0.515 & 0.337 \\ \midrule
VL-Time   & 0.876 & 0.438 & 0.509 & 0.645 & 0.479 \\ 
ByMyEyes  & 0.568 & 0.178 & 0.160 & 0.284 & 0.160 \\ \midrule
\ours{} & 0.899 & 0.515 & \textbf{0.627} & \textbf{0.751} & \textbf{0.568} \\ \bottomrule
\end{tabular}%
}
\caption{Comparison of marks-correct rate on the open-ended QA under all
scorers ($n{=}169$ items per method).} \label{tab:scorer_permethod}
\end{table}

\begin{table*}[t]
\centering
\resizebox{\textwidth}{!}{%
\begin{tabular}{@{}lccccccccc@{}}
\toprule
\textbf{Method} & \multicolumn{2}{c}{\textbf{\begin{tabular}[c]{@{}c@{}}Price Forecasting\\ (MAPE/MAE)\end{tabular}}} & \multicolumn{2}{c}{\textbf{\begin{tabular}[c]{@{}c@{}}Indicator Prediction\\ (MSE/MAE)\end{tabular}}}                                            & \multicolumn{2}{c}{\textbf{\begin{tabular}[c]{@{}c@{}}Trend Classification\\ (ACC/F1)\end{tabular}}}                                                 & \multicolumn{2}{c}{\textbf{\begin{tabular}[c]{@{}c@{}}Correlation Classification\\ (ACC/F1)\end{tabular}}} & \textbf{\begin{tabular}[c]{@{}c@{}}Question Answering\\ (ACC)\end{tabular}}       \\ 
\midrule
\midrule
Zero-Shot     & \begin{tabular}[c]{@{}c@{}}2.119\\ (±0.201)\end{tabular}          & \begin{tabular}[c]{@{}c@{}}1.806\\ (±0.426)\end{tabular}          & \begin{tabular}[c]{@{}c@{}}0.253\\ (±0.127)\end{tabular}          & \begin{tabular}[c]{@{}c@{}}0.214\\ (±0.025)\end{tabular}          & \begin{tabular}[c]{@{}c@{}}0.317\\ (±0.017)\end{tabular}          & \begin{tabular}[c]{@{}c@{}}0.389\\ (±0.018)\end{tabular}          & \begin{tabular}[c]{@{}c@{}}0.227\\ (±0.021)\end{tabular}          & \begin{tabular}[c]{@{}c@{}}0.189\\ (±0.028)\end{tabular}          & \begin{tabular}[c]{@{}c@{}}0.823\\ (±0.017)\end{tabular}          \\
+ MM  & \begin{tabular}[c]{@{}c@{}}2.050\\ (±0.129)\end{tabular}          & \begin{tabular}[c]{@{}c@{}}1.606\\ (±0.215)\end{tabular}          & \begin{tabular}[c]{@{}c@{}}0.860\\ (±0.658)\end{tabular}          & \begin{tabular}[c]{@{}c@{}}0.339\\ (±0.085)\end{tabular}          & \begin{tabular}[c]{@{}c@{}}0.437\\ (±0.033)\end{tabular}          & \begin{tabular}[c]{@{}c@{}}0.468\\ (±0.024)\end{tabular}          & \begin{tabular}[c]{@{}c@{}}0.313\\ (±0.031)\end{tabular}          & \begin{tabular}[c]{@{}c@{}}0.239\\ (±0.030)\end{tabular}          & \begin{tabular}[c]{@{}c@{}}0.847\\ (±0.012)\end{tabular}          \\
CoT           & \begin{tabular}[c]{@{}c@{}}2.217\\ (±0.147)\end{tabular}          & \begin{tabular}[c]{@{}c@{}}1.791\\ (±0.240)\end{tabular}          & \textbf{\begin{tabular}[c]{@{}c@{}}0.172\\ (±0.032)\end{tabular}} & \textbf{\begin{tabular}[c]{@{}c@{}}0.214\\ (±0.017)\end{tabular}} & \begin{tabular}[c]{@{}c@{}}0.310\\ (±0.014)\end{tabular}          & \begin{tabular}[c]{@{}c@{}}0.380\\ (±0.009)\end{tabular}          & \begin{tabular}[c]{@{}c@{}}0.240\\ (±0.016)\end{tabular}          & \begin{tabular}[c]{@{}c@{}}0.151\\ (±0.029)\end{tabular}          & \begin{tabular}[c]{@{}c@{}}0.797\\ (±0.026)\end{tabular}          \\
+ MM  & \textbf{\begin{tabular}[c]{@{}c@{}}1.912\\ (±0.079)\end{tabular}} & \begin{tabular}[c]{@{}c@{}}1.588\\ (±0.110)\end{tabular}          & \begin{tabular}[c]{@{}c@{}}2137.953\\ (±2789.155)\end{tabular}    & \begin{tabular}[c]{@{}c@{}}4.003\\ (±3.241)\end{tabular}          & \begin{tabular}[c]{@{}c@{}}0.437\\ (±0.021)\end{tabular}          & \begin{tabular}[c]{@{}c@{}}0.470\\ (±0.013)\end{tabular}          & \begin{tabular}[c]{@{}c@{}}0.303\\ (±0.012)\end{tabular}          & \begin{tabular}[c]{@{}c@{}}0.240\\ (±0.026)\end{tabular}          & \begin{tabular}[c]{@{}c@{}}0.817\\ (±0.025)\end{tabular}          \\
MAD           & \begin{tabular}[c]{@{}c@{}}2.362\\ (±0.036)\end{tabular}          & \begin{tabular}[c]{@{}c@{}}1.914\\ (±0.119)\end{tabular}          & \begin{tabular}[c]{@{}c@{}}2.043\\ (±2.679)\end{tabular}          & \begin{tabular}[c]{@{}c@{}}0.243\\ (±0.066)\end{tabular}          & \begin{tabular}[c]{@{}c@{}}0.353\\ (±0.025)\end{tabular}          & \begin{tabular}[c]{@{}c@{}}0.419\\ (±0.025)\end{tabular}          & \begin{tabular}[c]{@{}c@{}}0.247\\ (±0.021)\end{tabular}          & \begin{tabular}[c]{@{}c@{}}0.204\\ (±0.017)\end{tabular}          & \begin{tabular}[c]{@{}c@{}}0.750\\ (±0.008)\end{tabular}          \\
+ MM  & \begin{tabular}[c]{@{}c@{}}2.692\\ (±0.186)\end{tabular}          & \begin{tabular}[c]{@{}c@{}}2.535\\ (±0.610)\end{tabular}          & \begin{tabular}[c]{@{}c@{}}611.331\\ (±450.461)\end{tabular}      & \begin{tabular}[c]{@{}c@{}}5.567\\ (±2.484)\end{tabular}          & \begin{tabular}[c]{@{}c@{}}0.437\\ (±0.021)\end{tabular}          & \begin{tabular}[c]{@{}c@{}}0.465\\ (±0.032)\end{tabular}          & \begin{tabular}[c]{@{}c@{}}0.257\\ (±0.034)\end{tabular}          & \begin{tabular}[c]{@{}c@{}}0.236\\ (±0.053)\end{tabular}          & \begin{tabular}[c]{@{}c@{}}0.747\\ (±0.021)\end{tabular}          \\
\midrule
ByMyEyes      & \begin{tabular}[c]{@{}c@{}}219.122\\ (±65.619)\end{tabular}       & \begin{tabular}[c]{@{}c@{}}72.185\\ (±10.422)\end{tabular}        & \begin{tabular}[c]{@{}c@{}}1.054\\ (±0.151)\end{tabular}          & \begin{tabular}[c]{@{}c@{}}0.405\\ (±0.019)\end{tabular}          & \begin{tabular}[c]{@{}c@{}}0.640\\ (±0.022)\end{tabular}          & \begin{tabular}[c]{@{}c@{}}0.560\\ (±0.009)\end{tabular}          & \begin{tabular}[c]{@{}c@{}}0.250\\ (±0.014)\end{tabular}          & \begin{tabular}[c]{@{}c@{}}0.208\\ (±0.006)\end{tabular}          & \begin{tabular}[c]{@{}c@{}}0.823\\ (±0.046)\end{tabular}          \\
VL-Time       & \begin{tabular}[c]{@{}c@{}}2.526\\ (±0.522)\end{tabular}          & \begin{tabular}[c]{@{}c@{}}2.895\\ (±1.292)\end{tabular}          & \begin{tabular}[c]{@{}c@{}}0.282\\ (±0.036)\end{tabular}          & \begin{tabular}[c]{@{}c@{}}0.254\\ (±0.021)\end{tabular}          & \begin{tabular}[c]{@{}c@{}}0.330\\ (±0.028)\end{tabular}          & \begin{tabular}[c]{@{}c@{}}0.402\\ (±0.027)\end{tabular}          & \begin{tabular}[c]{@{}c@{}}0.270\\ (±0.022)\end{tabular}          & \begin{tabular}[c]{@{}c@{}}0.178\\ (±0.021)\end{tabular}          & \begin{tabular}[c]{@{}c@{}}0.773\\ (±0.026)\end{tabular}          \\
\midrule
\ours{}     & \begin{tabular}[c]{@{}c@{}}2.090\\ (±0.094)\end{tabular}          & \textbf{\begin{tabular}[c]{@{}c@{}}1.399\\ (±0.098)\end{tabular}} & \begin{tabular}[c]{@{}c@{}}0.176\\ (±0.053)\end{tabular}          & \begin{tabular}[c]{@{}c@{}}0.229\\ (±0.022)\end{tabular}          & \textbf{\begin{tabular}[c]{@{}c@{}}0.683\\ (±0.017)\end{tabular}} & \textbf{\begin{tabular}[c]{@{}c@{}}0.713\\ (±0.016)\end{tabular}} & \textbf{\begin{tabular}[c]{@{}c@{}}0.403\\ (±0.068)\end{tabular}} & \textbf{\begin{tabular}[c]{@{}c@{}}0.394\\ (±0.070)\end{tabular}} & \textbf{\begin{tabular}[c]{@{}c@{}}0.913\\ (±0.012)\end{tabular}} \\ \bottomrule
\end{tabular}%
}
\caption{Full experimental results on the \textbf{MTBench}\,(Finance) benchmark across different downstream tasks.} \label{table:mtbench_finance_full}
\end{table*}

\section{Extra Experimental Results} \label{section:extra_results}

This section reports full per-task results (mean$\pm$std over 3 runs) and additional analyses. Throughout, \textbf{lower is better} for error metrics (MAE/MAPE/MSE), while \textbf{higher is better} for ACC/F1/NLI-style scores. Beyond listing metrics, we provide concise interpretations to contextualize when and why \ours{} improves over baselines.

\subsection{MTBench}

\begin{table*}[t]
\centering
\resizebox{\textwidth}{!}{%
\begin{tabular}{@{}lccccccccccc@{}}
\toprule
\textbf{Method} & \multicolumn{2}{c}{\textbf{\begin{tabular}[c]{@{}c@{}}Forecasting\\ (MSE/MAE)\end{tabular}}}                                           & \multicolumn{2}{c}{\textbf{\begin{tabular}[c]{@{}c@{}}Indicator-Min\\ (MSE/MAE)\end{tabular}}}                                         & \multicolumn{2}{c}{\textbf{\begin{tabular}[c]{@{}c@{}}Indicator-Max\\ (MSE/MAE)\end{tabular}}}                                         & \multicolumn{2}{c}{\textbf{\begin{tabular}[c]{@{}c@{}}Indicator-Diff\\ (MSE/MAE)\end{tabular}}}                                        & \multicolumn{2}{c}{\textbf{\begin{tabular}[c]{@{}c@{}}Trend\\ (ACC/F1)\end{tabular}}} & \textbf{\begin{tabular}[c]{@{}c@{}}QA\\ (ACC)\end{tabular}}       \\ 
\midrule
\midrule
Zero-Shot     & \begin{tabular}[c]{@{}c@{}}26.474\\ (±1.841)\end{tabular}          & \begin{tabular}[c]{@{}c@{}}3.920\\ (±0.084)\end{tabular}          & \begin{tabular}[c]{@{}c@{}}19.054\\ (±2.116)\end{tabular}          & \begin{tabular}[c]{@{}c@{}}2.942\\ (±0.282)\end{tabular}          & \begin{tabular}[c]{@{}c@{}}52.335\\ (±12.056)\end{tabular}         & \begin{tabular}[c]{@{}c@{}}5.244\\ (±0.504)\end{tabular}          & \begin{tabular}[c]{@{}c@{}}42.132\\ (±8.913)\end{tabular}          & \begin{tabular}[c]{@{}c@{}}5.110\\ (±0.487)\end{tabular}          & \begin{tabular}[c]{@{}c@{}}0.250\\ (±0.008)\end{tabular}          & \begin{tabular}[c]{@{}c@{}}0.262\\ (±0.008)\end{tabular}          & \begin{tabular}[c]{@{}c@{}}0.617\\ (±0.048)\end{tabular}          \\
+ MM  & \begin{tabular}[c]{@{}c@{}}22.098\\ (±2.693)\end{tabular}          & \begin{tabular}[c]{@{}c@{}}3.531\\ (±0.186)\end{tabular}          & \textbf{\begin{tabular}[c]{@{}c@{}}14.990\\ (±1.924)\end{tabular}} & \begin{tabular}[c]{@{}c@{}}2.899\\ (±0.284)\end{tabular}          & \begin{tabular}[c]{@{}c@{}}34.086\\ (±4.292)\end{tabular}          & \begin{tabular}[c]{@{}c@{}}4.301\\ (±0.094)\end{tabular}          & \begin{tabular}[c]{@{}c@{}}33.297\\ (±4.591)\end{tabular}          & \begin{tabular}[c]{@{}c@{}}4.500\\ (±0.275)\end{tabular}          & \begin{tabular}[c]{@{}c@{}}0.523\\ (±0.021)\end{tabular}          & \begin{tabular}[c]{@{}c@{}}0.488\\ (±0.023)\end{tabular}          & \begin{tabular}[c]{@{}c@{}}0.653\\ (±0.026)\end{tabular}          \\
CoT           & \begin{tabular}[c]{@{}c@{}}28.967\\ (±2.942)\end{tabular}          & \begin{tabular}[c]{@{}c@{}}4.091\\ (±0.245)\end{tabular}          & \begin{tabular}[c]{@{}c@{}}16.568\\ (±3.070)\end{tabular}          & \begin{tabular}[c]{@{}c@{}}3.033\\ (±0.276)\end{tabular}          & \begin{tabular}[c]{@{}c@{}}23.541\\ (±2.822)\end{tabular}          & \textbf{\begin{tabular}[c]{@{}c@{}}3.507\\ (±0.141)\end{tabular}} & \begin{tabular}[c]{@{}c@{}}23.335\\ (±3.221)\end{tabular}          & \begin{tabular}[c]{@{}c@{}}3.530\\ (±0.162)\end{tabular}          & \begin{tabular}[c]{@{}c@{}}0.220\\ (±0.036)\end{tabular}          & \begin{tabular}[c]{@{}c@{}}0.234\\ (±0.040)\end{tabular}          & \begin{tabular}[c]{@{}c@{}}0.600\\ (±0.037)\end{tabular}          \\
+ MM  & \textbf{\begin{tabular}[c]{@{}c@{}}22.005\\ (±1.645)\end{tabular}} & \textbf{\begin{tabular}[c]{@{}c@{}}3.523\\ (±0.114)\end{tabular}} & \begin{tabular}[c]{@{}c@{}}16.025\\ (±3.045)\end{tabular}          & \textbf{\begin{tabular}[c]{@{}c@{}}2.890\\ (±0.382)\end{tabular}} & \begin{tabular}[c]{@{}c@{}}35.665\\ (±3.367)\end{tabular}          & \begin{tabular}[c]{@{}c@{}}4.342\\ (±0.096)\end{tabular}          & \begin{tabular}[c]{@{}c@{}}36.671\\ (±2.025)\end{tabular}          & \begin{tabular}[c]{@{}c@{}}4.682\\ (±0.183)\end{tabular}          & \begin{tabular}[c]{@{}c@{}}0.477\\ (±0.039)\end{tabular}          & \begin{tabular}[c]{@{}c@{}}0.457\\ (±0.036)\end{tabular}          & \begin{tabular}[c]{@{}c@{}}0.617\\ (±0.021)\end{tabular}          \\
MAD           & \begin{tabular}[c]{@{}c@{}}27.051\\ (±2.202)\end{tabular}          & \begin{tabular}[c]{@{}c@{}}3.958\\ (±0.135)\end{tabular}          & \begin{tabular}[c]{@{}c@{}}17.327\\ (±4.716)\end{tabular}          & \begin{tabular}[c]{@{}c@{}}3.035\\ (±0.360)\end{tabular}          & \begin{tabular}[c]{@{}c@{}}39.936\\ (±11.614)\end{tabular}         & \begin{tabular}[c]{@{}c@{}}4.706\\ (±0.497)\end{tabular}          & \begin{tabular}[c]{@{}c@{}}35.600\\ (±2.759)\end{tabular}          & \begin{tabular}[c]{@{}c@{}}4.707\\ (±0.156)\end{tabular}          & \begin{tabular}[c]{@{}c@{}}0.200\\ (±0.014)\end{tabular}          & \begin{tabular}[c]{@{}c@{}}0.175\\ (±0.024)\end{tabular}          & \begin{tabular}[c]{@{}c@{}}0.520\\ (±0.036)\end{tabular}          \\
+ MM  & \begin{tabular}[c]{@{}c@{}}27.594\\ (±1.826)\end{tabular}          & \begin{tabular}[c]{@{}c@{}}3.965\\ (±0.204)\end{tabular}          & \begin{tabular}[c]{@{}c@{}}16.126\\ (±3.898)\end{tabular}          & \begin{tabular}[c]{@{}c@{}}2.973\\ (±0.331)\end{tabular}          & \begin{tabular}[c]{@{}c@{}}26.999\\ (±7.123)\end{tabular}          & \begin{tabular}[c]{@{}c@{}}3.804\\ (±0.422)\end{tabular}          & \begin{tabular}[c]{@{}c@{}}28.924\\ (±2.038)\end{tabular}          & \begin{tabular}[c]{@{}c@{}}4.182\\ (±0.207)\end{tabular}          & \begin{tabular}[c]{@{}c@{}}0.320\\ (±0.029)\end{tabular}          & \begin{tabular}[c]{@{}c@{}}0.347\\ (±0.024)\end{tabular}          & \begin{tabular}[c]{@{}c@{}}0.500\\ (±0.022)\end{tabular}          \\
\midrule
ByMyEyes      & \begin{tabular}[c]{@{}c@{}}39.632\\ (±5.722)\end{tabular}          & \begin{tabular}[c]{@{}c@{}}4.714\\ (±0.252)\end{tabular}          & \begin{tabular}[c]{@{}c@{}}23.296\\ (±7.799)\end{tabular}          & \begin{tabular}[c]{@{}c@{}}3.224\\ (±0.434)\end{tabular}          & \begin{tabular}[c]{@{}c@{}}51.906\\ (±6.633)\end{tabular}          & \begin{tabular}[c]{@{}c@{}}5.127\\ (±0.460)\end{tabular}          & \begin{tabular}[c]{@{}c@{}}51.179\\ (±3.751)\end{tabular}          & \begin{tabular}[c]{@{}c@{}}5.887\\ (±0.249)\end{tabular}          & \begin{tabular}[c]{@{}c@{}}0.190\\ (±0.008)\end{tabular}          & \begin{tabular}[c]{@{}c@{}}0.167\\ (±0.016)\end{tabular}          & \begin{tabular}[c]{@{}c@{}}0.593\\ (±0.031)\end{tabular}          \\
VL-Time       & \begin{tabular}[c]{@{}c@{}}37.258\\ (±3.781)\end{tabular}          & \begin{tabular}[c]{@{}c@{}}4.695\\ (±0.180)\end{tabular}          & \begin{tabular}[c]{@{}c@{}}15.293\\ (±1.605)\end{tabular}          & \begin{tabular}[c]{@{}c@{}}2.966\\ (±0.197)\end{tabular}          & \textbf{\begin{tabular}[c]{@{}c@{}}22.369\\ (±2.675)\end{tabular}} & \begin{tabular}[c]{@{}c@{}}3.528\\ (±0.210)\end{tabular}          & \begin{tabular}[c]{@{}c@{}}19.495\\ (±0.607)\end{tabular}          & \begin{tabular}[c]{@{}c@{}}3.337\\ (±0.084)\end{tabular}          & \begin{tabular}[c]{@{}c@{}}0.220\\ (±0.029)\end{tabular}          & \begin{tabular}[c]{@{}c@{}}0.198\\ (±0.013)\end{tabular}          & \begin{tabular}[c]{@{}c@{}}0.613\\ (±0.039)\end{tabular}          \\
\midrule
\ours{}     & \begin{tabular}[c]{@{}c@{}}29.633\\ (±2.721)\end{tabular}          & \begin{tabular}[c]{@{}c@{}}4.162\\ (±0.244)\end{tabular}          & \begin{tabular}[c]{@{}c@{}}36.179\\ (±7.078)\end{tabular}          & \begin{tabular}[c]{@{}c@{}}3.192\\ (±0.363)\end{tabular}          & \begin{tabular}[c]{@{}c@{}}52.580\\ (±12.187)\end{tabular}         & \begin{tabular}[c]{@{}c@{}}3.522\\ (±0.391)\end{tabular}          & \textbf{\begin{tabular}[c]{@{}c@{}}12.197\\ (±2.366)\end{tabular}} & \textbf{\begin{tabular}[c]{@{}c@{}}2.334\\ (±0.273)\end{tabular}} & \textbf{\begin{tabular}[c]{@{}c@{}}0.557\\ (±0.025)\end{tabular}} & \textbf{\begin{tabular}[c]{@{}c@{}}0.583\\ (±0.016)\end{tabular}} & \textbf{\begin{tabular}[c]{@{}c@{}}0.717\\ (±0.052)\end{tabular}} \\ \bottomrule
\end{tabular}%
}
\caption{Full experimental results on the \textbf{MTBench}\,(Weather) benchmark across different downstream tasks.} \label{table:mtbench_weather_full}
\end{table*}

\subsubsection{Finance}

\autoref{table:mtbench_finance_full} reports results for price forecasting (MAPE/MAE), indicator prediction (MSE/MAE), trend classification (ACC/F1), news correlation classification (ACC/F1), and financial multi-choice question answering (ACC). Overall, \ours{} delivers its largest benefits on \emph{discrete reasoning} tasks that require aggregating evidence across views and validating consistency.

\paragraph{Forecasting and indicator prediction} On price forecasting, \ours{} attains the best MAE ($1.399$), improving over the strongest baseline MAE ($1.588$) by $\approx 11.9\%$ relative reduction. However, \ours{} is not the best on MAPE ($2.090$ vs.\ $1.912$ for the best baseline), suggesting that normalized percentage errors can remain sensitive to scaling/denominator effects even when absolute error is reduced.
For indicator prediction, \ours{} remains competitive (MSE/MAE $0.176/0.229$), close to the best single-agent CoT baseline ($0.172/0.214$). Notably, na\"ive multi-modal extensions of debate-style baselines can become numerically unstable on this regression-style task (e.g., extremely large MSE values), highlighting that simply adding visual inputs does not reliably yield calibrated numerical reasoning. In contrast, \ours{} avoids catastrophic failures and maintains low variance, but closing the remaining gap to the best indicator-prediction baseline is a clear opportunity for improvement.

\paragraph{Trend, correlation, and QA} \ours{} achieves the best performance on all three discrete finance tasks. On trend classification, \ours{} reaches $0.683/0.713$ (ACC/F1), improving over the strongest baseline ($0.640/0.560$) by $+4.3$ ACC points and $+15.3$ F1 points. On correlation classification, \ours{} attains $0.403/0.394$ (ACC/F1), outperforming the best baseline (around $0.313/0.240$) by $+9.0$ ACC points and $+15.4$ F1 points. For MCQA, \ours{} achieves $0.913$ ACC, exceeding the best baseline ($0.847$) by $+6.6$ points. These gains are consistent with the core design of \ours{}: combining frequency-domain cues with cross-view verification reduces spurious pattern matching and improves decision stability on label-centric tasks.

\begin{table*}[t]
\centering
\resizebox{\textwidth}{!}{%
\begin{tabular}{@{}lcccccc|cccccc@{}}
\toprule
\multirow{3}{*}{\textbf{Method}} & \multicolumn{6}{c}{\textbf{Accuracy}} & \multicolumn{6}{|c}{\textbf{F1}} \\ \cmidrule(l){2-13} 
 & \multicolumn{2}{c}{\textbf{Simple Deterministic}} & \multicolumn{2}{c}{\textbf{Complex Deterministic}} & \multicolumn{2}{c}{\textbf{Probabilistic}} & \multicolumn{2}{|c}{\textbf{Simple Deterministic}} & \multicolumn{2}{c}{\textbf{Complex Deterministic}} & \multicolumn{2}{c}{\textbf{Probabilistic}} \\
                                 & \textbf{RCW}                                                      & \textbf{TEE}                                                      & \textbf{ECG}                                                      & \textbf{EMG}                                                      & \textbf{CTU}                                                      & \textbf{HAR}                                                      & \textbf{RCW}                                                      & \textbf{TEE}                                                      & \textbf{ECG}                                                      & \textbf{EMG}                                                      & \textbf{CTU}                                                      & \textbf{HAR}                                                      \\ 
\midrule
\midrule
Zero-Shot                        & \begin{tabular}[c]{@{}c@{}}0.470\\ (±0.029)\end{tabular}          & \begin{tabular}[c]{@{}c@{}}0.128\\ (±0.017)\end{tabular}          & \begin{tabular}[c]{@{}c@{}}0.220\\ (±0.016)\end{tabular}          & \begin{tabular}[c]{@{}c@{}}0.154\\ (±0.011)\end{tabular}          & \begin{tabular}[c]{@{}c@{}}0.507\\ (±0.025)\end{tabular}          & \begin{tabular}[c]{@{}c@{}}0.237\\ (±0.005)\end{tabular}          & \begin{tabular}[c]{@{}c@{}}0.319\\ (±0.014)\end{tabular}          & \begin{tabular}[c]{@{}c@{}}0.059\\ (±0.006)\end{tabular}          & \begin{tabular}[c]{@{}c@{}}0.116\\ (±0.001)\end{tabular}          & \begin{tabular}[c]{@{}c@{}}0.088\\ (±0.020)\end{tabular}          & \begin{tabular}[c]{@{}c@{}}0.373\\ (±0.040)\end{tabular}          & \begin{tabular}[c]{@{}c@{}}0.123\\ (±0.006)\end{tabular}          \\
+ MM                     & \begin{tabular}[c]{@{}c@{}}0.483\\ (±0.026)\end{tabular}          & \begin{tabular}[c]{@{}c@{}}0.187\\ (±0.013)\end{tabular}          & \begin{tabular}[c]{@{}c@{}}0.250\\ (±0.008)\end{tabular}          & \begin{tabular}[c]{@{}c@{}}0.236\\ (±0.023)\end{tabular}          & \begin{tabular}[c]{@{}c@{}}0.503\\ (±0.012)\end{tabular}          & \begin{tabular}[c]{@{}c@{}}0.243\\ (±0.026)\end{tabular}          & \begin{tabular}[c]{@{}c@{}}0.408\\ (±0.022)\end{tabular}          & \begin{tabular}[c]{@{}c@{}}0.124\\ (±0.017)\end{tabular}          & \begin{tabular}[c]{@{}c@{}}0.107\\ (±0.011)\end{tabular}          & \begin{tabular}[c]{@{}c@{}}0.221\\ (±0.024)\end{tabular}          & \begin{tabular}[c]{@{}c@{}}0.480\\ (±0.014)\end{tabular}          & \begin{tabular}[c]{@{}c@{}}0.160\\ (±0.016)\end{tabular}          \\
CoT                              & \begin{tabular}[c]{@{}c@{}}0.497\\ (±0.005)\end{tabular}          & \begin{tabular}[c]{@{}c@{}}0.137\\ (±0.034)\end{tabular}          & \begin{tabular}[c]{@{}c@{}}0.187\\ (±0.026)\end{tabular}          & \begin{tabular}[c]{@{}c@{}}0.455\\ (±0.041)\end{tabular}          & \begin{tabular}[c]{@{}c@{}}0.483\\ (±0.005)\end{tabular}          & \begin{tabular}[c]{@{}c@{}}0.297\\ (±0.009)\end{tabular}          & \begin{tabular}[c]{@{}c@{}}0.338\\ (±0.009)\end{tabular}          & \begin{tabular}[c]{@{}c@{}}0.083\\ (±0.021)\end{tabular}          & \begin{tabular}[c]{@{}c@{}}0.125\\ (±0.016)\end{tabular}          & \begin{tabular}[c]{@{}c@{}}0.457\\ (±0.041)\end{tabular}          & \begin{tabular}[c]{@{}c@{}}0.398\\ (±0.012)\end{tabular}          & \begin{tabular}[c]{@{}c@{}}0.158\\ (±0.008)\end{tabular}          \\
+ MM                     & \begin{tabular}[c]{@{}c@{}}0.477\\ (±0.017)\end{tabular}          & \begin{tabular}[c]{@{}c@{}}0.247\\ (±0.030)\end{tabular}          & \begin{tabular}[c]{@{}c@{}}0.247\\ (±0.012)\end{tabular}          & \begin{tabular}[c]{@{}c@{}}0.203\\ (±0.046)\end{tabular}          & \begin{tabular}[c]{@{}c@{}}0.500\\ (±0.008)\end{tabular}          & \begin{tabular}[c]{@{}c@{}}0.317\\ (±0.025)\end{tabular}          & \begin{tabular}[c]{@{}c@{}}0.390\\ (±0.011)\end{tabular}          & \begin{tabular}[c]{@{}c@{}}0.160\\ (±0.027)\end{tabular}          & \begin{tabular}[c]{@{}c@{}}0.111\\ (±0.011)\end{tabular}          & \begin{tabular}[c]{@{}c@{}}0.194\\ (±0.085)\end{tabular}          & \begin{tabular}[c]{@{}c@{}}0.445\\ (±0.014)\end{tabular}          & \begin{tabular}[c]{@{}c@{}}0.196\\ (±0.018)\end{tabular}          \\
MAD                              & \begin{tabular}[c]{@{}c@{}}0.490\\ (±0.008)\end{tabular}          & \begin{tabular}[c]{@{}c@{}}0.151\\ (±0.022)\end{tabular}          & \begin{tabular}[c]{@{}c@{}}0.167\\ (±0.012)\end{tabular}          & \begin{tabular}[c]{@{}c@{}}0.171\\ (±0.020)\end{tabular}          & \begin{tabular}[c]{@{}c@{}}0.517\\ (±0.005)\end{tabular}          & \begin{tabular}[c]{@{}c@{}}0.267\\ (±0.017)\end{tabular}          & \begin{tabular}[c]{@{}c@{}}0.329\\ (±0.004)\end{tabular}          & \begin{tabular}[c]{@{}c@{}}0.069\\ (±0.015)\end{tabular}          & \begin{tabular}[c]{@{}c@{}}0.110\\ (±0.008)\end{tabular}          & \begin{tabular}[c]{@{}c@{}}0.124\\ (±0.022)\end{tabular}          & \begin{tabular}[c]{@{}c@{}}0.430\\ (±0.007)\end{tabular}          & \begin{tabular}[c]{@{}c@{}}0.178\\ (±0.015)\end{tabular}          \\
+ MM                     & \begin{tabular}[c]{@{}c@{}}0.503\\ (±0.005)\end{tabular}          & \begin{tabular}[c]{@{}c@{}}0.178\\ (±0.000)\end{tabular}          & \begin{tabular}[c]{@{}c@{}}0.240\\ (±0.016)\end{tabular}          & \begin{tabular}[c]{@{}c@{}}0.203\\ (±0.064)\end{tabular}          & \begin{tabular}[c]{@{}c@{}}0.523\\ (±0.017)\end{tabular}          & \begin{tabular}[c]{@{}c@{}}0.253\\ (±0.040)\end{tabular}          & \begin{tabular}[c]{@{}c@{}}0.386\\ (±0.019)\end{tabular}          & \begin{tabular}[c]{@{}c@{}}0.098\\ (±0.017)\end{tabular}          & \begin{tabular}[c]{@{}c@{}}0.212\\ (±0.020)\end{tabular}          & \begin{tabular}[c]{@{}c@{}}0.230\\ (±0.065)\end{tabular}          & \textbf{\begin{tabular}[c]{@{}c@{}}0.519\\ (±0.016)\end{tabular}} & \begin{tabular}[c]{@{}c@{}}0.182\\ (±0.037)\end{tabular}          \\
\midrule
ByMyEyes                         & \begin{tabular}[c]{@{}c@{}}0.463\\ (±0.034)\end{tabular}          & \begin{tabular}[c]{@{}c@{}}0.201\\ (±0.045)\end{tabular}          & \begin{tabular}[c]{@{}c@{}}0.267\\ (±0.017)\end{tabular}          & \begin{tabular}[c]{@{}c@{}}0.122\\ (±0.034)\end{tabular}          & \begin{tabular}[c]{@{}c@{}}0.467\\ (±0.033)\end{tabular}          & \begin{tabular}[c]{@{}c@{}}0.220\\ (±0.028)\end{tabular}          & \begin{tabular}[c]{@{}c@{}}0.385\\ (±0.034)\end{tabular}          & \begin{tabular}[c]{@{}c@{}}0.168\\ (±0.040)\end{tabular}          & \begin{tabular}[c]{@{}c@{}}0.158\\ (±0.036)\end{tabular}          & \begin{tabular}[c]{@{}c@{}}0.055\\ (±0.045)\end{tabular}          & \begin{tabular}[c]{@{}c@{}}0.343\\ (±0.023)\end{tabular}          & \begin{tabular}[c]{@{}c@{}}0.174\\ (±0.028)\end{tabular}          \\
VL-Time                          & \begin{tabular}[c]{@{}c@{}}0.463\\ (±0.009)\end{tabular}          & \begin{tabular}[c]{@{}c@{}}0.260\\ (±0.051)\end{tabular}          & \begin{tabular}[c]{@{}c@{}}0.260\\ (±0.008)\end{tabular}          & \begin{tabular}[c]{@{}c@{}}0.203\\ (±0.057)\end{tabular}          & \begin{tabular}[c]{@{}c@{}}0.493\\ (±0.005)\end{tabular}          & \begin{tabular}[c]{@{}c@{}}0.313\\ (±0.012)\end{tabular}          & \begin{tabular}[c]{@{}c@{}}0.438\\ (±0.014)\end{tabular}          & \begin{tabular}[c]{@{}c@{}}0.214\\ (±0.039)\end{tabular}          & \begin{tabular}[c]{@{}c@{}}0.131\\ (±0.021)\end{tabular}          & \begin{tabular}[c]{@{}c@{}}0.187\\ (±0.055)\end{tabular}          & \begin{tabular}[c]{@{}c@{}}0.383\\ (±0.023)\end{tabular}          & \begin{tabular}[c]{@{}c@{}}0.198\\ (±0.021)\end{tabular}          \\
\midrule
\ours{}                        & \textbf{\begin{tabular}[c]{@{}c@{}}0.550\\ (±0.008)\end{tabular}} & \textbf{\begin{tabular}[c]{@{}c@{}}0.288\\ (±0.039)\end{tabular}} & \textbf{\begin{tabular}[c]{@{}c@{}}0.307\\ (±0.012)\end{tabular}} & \textbf{\begin{tabular}[c]{@{}c@{}}0.488\\ (±0.159)\end{tabular}} & \textbf{\begin{tabular}[c]{@{}c@{}}0.573\\ (±0.045)\end{tabular}} & \textbf{\begin{tabular}[c]{@{}c@{}}0.323\\ (±0.041)\end{tabular}} & \textbf{\begin{tabular}[c]{@{}c@{}}0.440\\ (±0.017)\end{tabular}} & \textbf{\begin{tabular}[c]{@{}c@{}}0.241\\ (±0.041)\end{tabular}} & \textbf{\begin{tabular}[c]{@{}c@{}}0.281\\ (±0.019)\end{tabular}} & \textbf{\begin{tabular}[c]{@{}c@{}}0.491\\ (±0.168)\end{tabular}} & \begin{tabular}[c]{@{}c@{}}0.473\\ (±0.080)\end{tabular}          & \textbf{\begin{tabular}[c]{@{}c@{}}0.240\\ (±0.028)\end{tabular}} \\ \bottomrule
\end{tabular}%
}
\caption{Full experimental results on the \textbf{TimerBed} benchmark across reasoning patterns and classification tasks.} \label{table:timerbed_full}
\vspace{-0.3cm}
\end{table*}

\subsubsection{Weather}
\autoref{table:mtbench_weather_full} reports results for temperature forecasting, min, max, and difference temperature's indicator prediction (MSE/MAE), trend classification (ACC/F1), and multi-choice weather question answering (ACC). In contrast to finance, weather results reveal a more nuanced picture: \ours{} is strongest on \emph{structural} and \emph{relative} reasoning, but does not dominate point-wise regression.

\paragraph{Relative reasoning and discrete decisions} \ours{} achieves the best performance on (i) \textbf{Indicator-Diff}, (ii) trend classification, and (iii) QA. For Indicator-Diff, \ours{} reduces MSE from $19.495$ to $12.197$ ($\approx 37.4\%$ reduction) and reduces MAE from $3.337$ to $2.334$ ($\approx 30.1\%$ reduction). This pattern aligns with our hypothesis that frequency-domain evidence and cross-panel checks are especially effective when the target is a \emph{difference/derivative}-like quantity, where global structure is more informative than absolute level. Similarly, \ours{} improves trend ACC/F1 to $0.557/0.583$ (vs.\ $0.523/0.488$ for the best baseline) and boosts QA accuracy to $0.717$ (vs.\ $0.653$).

\paragraph{Pointwise forecasting and absolute extrema} For forecasting and absolute extrema (Indicator-Min/Max), \ours{} underperforms the best baselines. In particular, forecasting MSE/MAE for \ours{} is $29.633/4.162$, compared to the best baseline at $22.005/3.523$. A similar gap appears for Indicator-Min and Indicator-Max in MSE. These results suggest that while \ours{} improves reliability for \emph{comparative} judgments (trend/QA) and \emph{relative} indicators (diff), it can be less effective for tasks requiring precise reconstruction of absolute values where strong, direct visual pattern extraction (or simpler multimodal prompting) may already suffice. This points to a practical limitation: coupling the verifier with stronger value-calibrated regression priors may be necessary for weather-style point forecasting.

\begin{table*}[t]
\centering
\resizebox{\textwidth}{!}{%
\begin{tabular}{@{}lcccccc|ccc@{}}
\toprule
\multirow{2}{*}{\textbf{Method}} & \multicolumn{6}{c}{\textbf{Classical Numerical Task}} & \multicolumn{3}{|c}{\textbf{Question Answering}} \\ \cmidrule(l){2-10} 
                             & \multicolumn{2}{c}{\textbf{\begin{tabular}[c]{@{}c@{}}Forecasting\\ (MAE/MAPE)\end{tabular}}}                                            & \multicolumn{2}{c}{\textbf{\begin{tabular}[c]{@{}c@{}}Imputation\\ (MAE/MAPE)\end{tabular}}}                                           & \textbf{\begin{tabular}[c]{@{}c@{}}Anomaly\\ (ACC)\end{tabular}}  & \textbf{\begin{tabular}[c]{@{}c@{}}Classification\\ (ACC)\end{tabular}} & \textbf{\begin{tabular}[c]{@{}c@{}}Judgment\\ (ACC)\end{tabular}} & \textbf{\begin{tabular}[c]{@{}c@{}}Multiple Choice\\ (ACC)\end{tabular}} & \textbf{\begin{tabular}[c]{@{}c@{}}Open-Ended \\ (NLI)\end{tabular}} \\
\midrule
\midrule
Zero-Shot                        & \begin{tabular}[c]{@{}c@{}}95.929\\ (±52.292)\end{tabular}          & \begin{tabular}[c]{@{}c@{}}321.737\\ (±196.751)\end{tabular}       & \begin{tabular}[c]{@{}c@{}}96.854\\ (±123.464)\end{tabular}       & \begin{tabular}[c]{@{}c@{}}13.405\\ (±7.713)\end{tabular}          & \begin{tabular}[c]{@{}c@{}}0.543\\ (±0.062)\end{tabular}          & \begin{tabular}[c]{@{}c@{}}0.213\\ (±0.029)\end{tabular}                & \begin{tabular}[c]{@{}c@{}}0.772\\ (±0.085)\end{tabular}          & \begin{tabular}[c]{@{}c@{}}0.406\\ (±0.070)\end{tabular}                 & \begin{tabular}[c]{@{}c@{}}0.875\\ (±0.031)\end{tabular}             \\
+ MM                     & \begin{tabular}[c]{@{}c@{}}88.319\\ (±52.244)\end{tabular}          & \begin{tabular}[c]{@{}c@{}}400.442\\ (±293.317)\end{tabular}       & \textbf{\begin{tabular}[c]{@{}c@{}}6.729\\ (±1.122)\end{tabular}} & \textbf{\begin{tabular}[c]{@{}c@{}}10.758\\ (±5.643)\end{tabular}} & \begin{tabular}[c]{@{}c@{}}0.583\\ (±0.012)\end{tabular}          & \begin{tabular}[c]{@{}c@{}}0.243\\ (±0.025)\end{tabular}                & \begin{tabular}[c]{@{}c@{}}0.650\\ (±0.024)\end{tabular}          & \begin{tabular}[c]{@{}c@{}}0.536\\ (±0.066)\end{tabular}                 & \begin{tabular}[c]{@{}c@{}}0.749\\ (±0.057)\end{tabular}             \\
CoT                              & \begin{tabular}[c]{@{}c@{}}105.363\\ (±69.763)\end{tabular}         & \begin{tabular}[c]{@{}c@{}}496.511\\ (±457.050)\end{tabular}       & \begin{tabular}[c]{@{}c@{}}33.569\\ (±28.374)\end{tabular}        & \begin{tabular}[c]{@{}c@{}}18.654\\ (±9.779)\end{tabular}          & \begin{tabular}[c]{@{}c@{}}0.583\\ (±0.041)\end{tabular}          & \begin{tabular}[c]{@{}c@{}}0.507\\ (±0.026)\end{tabular}                & \begin{tabular}[c]{@{}c@{}}0.826\\ (±0.019)\end{tabular}          & \begin{tabular}[c]{@{}c@{}}0.360\\ (±0.078)\end{tabular}                 & \begin{tabular}[c]{@{}c@{}}0.858\\ (±0.018)\end{tabular}             \\
+ MM                     & \begin{tabular}[c]{@{}c@{}}91.677\\ (±63.708)\end{tabular}          & \begin{tabular}[c]{@{}c@{}}422.630\\ (±206.218)\end{tabular}       & \begin{tabular}[c]{@{}c@{}}11.058\\ (±5.078)\end{tabular}         & \begin{tabular}[c]{@{}c@{}}11.711\\ (±5.684)\end{tabular}          & \begin{tabular}[c]{@{}c@{}}0.533\\ (±0.042)\end{tabular}          & \begin{tabular}[c]{@{}c@{}}0.513\\ (±0.040)\end{tabular}                & \begin{tabular}[c]{@{}c@{}}0.713\\ (±0.076)\end{tabular}          & \begin{tabular}[c]{@{}c@{}}0.322\\ (±0.047)\end{tabular}                 & \textbf{\begin{tabular}[c]{@{}c@{}}0.911\\ (±0.016)\end{tabular}}    \\
MAD                              & \begin{tabular}[c]{@{}c@{}}131.493\\ (±60.633)\end{tabular}         & \begin{tabular}[c]{@{}c@{}}241.480\\ (±96.291)\end{tabular}        & \begin{tabular}[c]{@{}c@{}}30.700\\ (±20.733)\end{tabular}        & \begin{tabular}[c]{@{}c@{}}34.614\\ (±34.713)\end{tabular}         & \begin{tabular}[c]{@{}c@{}}0.577\\ (±0.029)\end{tabular}          & \begin{tabular}[c]{@{}c@{}}0.553\\ (±0.052)\end{tabular}                & \begin{tabular}[c]{@{}c@{}}0.845\\ (±0.068)\end{tabular}          & \begin{tabular}[c]{@{}c@{}}0.631\\ (±0.062)\end{tabular}                 & \begin{tabular}[c]{@{}c@{}}0.803\\ (±0.056)\end{tabular}             \\
+ MM                     & \begin{tabular}[c]{@{}c@{}}144.476\\ (±64.677)\end{tabular}         & \begin{tabular}[c]{@{}c@{}}173.558\\ (±61.746)\end{tabular}        & \begin{tabular}[c]{@{}c@{}}27.302\\ (±28.741)\end{tabular}        & \begin{tabular}[c]{@{}c@{}}12.063\\ (±2.320)\end{tabular}          & \begin{tabular}[c]{@{}c@{}}0.543\\ (±0.038)\end{tabular}          & \begin{tabular}[c]{@{}c@{}}0.540\\ (±0.045)\end{tabular}                & \begin{tabular}[c]{@{}c@{}}0.679\\ (±0.152)\end{tabular}          & \begin{tabular}[c]{@{}c@{}}0.579\\ (±0.112)\end{tabular}                 & \begin{tabular}[c]{@{}c@{}}0.750\\ (±0.058)\end{tabular}             \\ \midrule
ByMyEyes                         & \begin{tabular}[c]{@{}c@{}}318.166\\ (±166.606)\end{tabular}        & \begin{tabular}[c]{@{}c@{}}411.087\\ (±119.951)\end{tabular}       & \begin{tabular}[c]{@{}c@{}}588.589\\ (±76.385)\end{tabular}       & \begin{tabular}[c]{@{}c@{}}522.526\\ (±537.189)\end{tabular}       & \begin{tabular}[c]{@{}c@{}}0.523\\ (±0.049)\end{tabular}          & \begin{tabular}[c]{@{}c@{}}0.243\\ (±0.083)\end{tabular}                & \begin{tabular}[c]{@{}c@{}}0.627\\ (±0.114)\end{tabular}          & \begin{tabular}[c]{@{}c@{}}0.579\\ (±0.029)\end{tabular}                 & \begin{tabular}[c]{@{}c@{}}0.565\\ (±0.110)\end{tabular}             \\
VL-Time                          & \begin{tabular}[c]{@{}c@{}}97.402\\ (±63.517)\end{tabular}          & \begin{tabular}[c]{@{}c@{}}258.042\\ (±96.573)\end{tabular}        & \begin{tabular}[c]{@{}c@{}}25.186\\ (±1.501)\end{tabular}         & \begin{tabular}[c]{@{}c@{}}15.066\\ (±2.116)\end{tabular}          & \begin{tabular}[c]{@{}c@{}}0.527\\ (±0.062)\end{tabular}          & \begin{tabular}[c]{@{}c@{}}0.470\\ (±0.036)\end{tabular}                & \begin{tabular}[c]{@{}c@{}}0.480\\ (±0.132)\end{tabular}          & \begin{tabular}[c]{@{}c@{}}0.393\\ (±0.076)\end{tabular}                 & \begin{tabular}[c]{@{}c@{}}0.877\\ (±0.035)\end{tabular}             \\
\midrule
\ours{}                  & \textbf{\begin{tabular}[c]{@{}c@{}}37.346\\ (±11.828)\end{tabular}} & \textbf{\begin{tabular}[c]{@{}c@{}}72.308\\ (±8.686)\end{tabular}} & \begin{tabular}[c]{@{}c@{}}18.343\\ (±7.039)\end{tabular}         & \begin{tabular}[c]{@{}c@{}}51.395\\ (±3.151)\end{tabular}          & \textbf{\begin{tabular}[c]{@{}c@{}}0.623\\ (±0.025)\end{tabular}} & \textbf{\begin{tabular}[c]{@{}c@{}}0.607\\ (±0.021)\end{tabular}}       & \textbf{\begin{tabular}[c]{@{}c@{}}0.897\\ (±0.045)\end{tabular}} & \textbf{\begin{tabular}[c]{@{}c@{}}0.796\\ (±0.101)\end{tabular}}        & \begin{tabular}[c]{@{}c@{}}0.899\\ (±0.045)\end{tabular}             \\ \bottomrule
\end{tabular}%
}
\caption{Full experimental results on the \textbf{TSQA} benchmark across different downstream tasks.} \label{table:tsqa_full}
\end{table*}

\subsection{TimerBed}

\autoref{table:timerbed_full} reports results on TimerBed across six classification datasets (RCW, TEE, ECG, EMG, CTU, HAR), grouped by reasoning pattern (simple/complex deterministic and probabilistic). TimerBed is particularly diagnostic for whether a method can maintain consistent decision logic under varying temporal regularities.

\paragraph{Overall performance} \ours{} attains the best accuracy on all six tasks and the best F1 on five out of six tasks. The gains are most pronounced on the complex deterministic datasets. \ours{} improves ECG accuracy to $0.307$ (vs.\ $0.267$) and ECG F1 to $0.281$ (vs.\ $0.212$), indicating that multi-step temporal reasoning and verification are beneficial when the decision boundary depends on subtle, non-local temporal structure.

\paragraph{Calibration trade-off on CTU (probabilistic)} On CTU, \ours{} achieves the highest accuracy ($0.573$) but not the highest F1 ($0.473$ vs.\ $0.519$ for the best baseline). This discrepancy is consistent with a threshold/calibration trade-off under class imbalance: a method can optimize accuracy by favoring the majority class while sacrificing minority-class recall, which impacts F1 more directly. Improving probability calibration or incorporating class-sensitive verification criteria could close this gap without sacrificing overall accuracy.

\subsection{TSQA}
\autoref{table:tsqa_full} provides full results on TSQA, covering classical numerical tasks (forecasting, imputation, anomaly detection, classification) and three QA formats (judgment, multiple choice, open-ended questions). Overall, TSQA highlights a clear strength of \ours{} on forecasting and discrete decision tasks, alongside a notable weakness on imputation.

\subsubsection{Forecasting}
\ours{} achieves the best forecasting performance by a large margin: MAE/MAPE of $37.346/72.308$, reducing error relative to the strongest baseline by $\approx 57.7\%$ (MAE) and $\approx 58.3\%$ (MAPE). Beyond mean improvements, \ours{} also substantially reduces run-to-run variance (e.g., forecasting std drops from tens to low double digits), suggesting that verifier-based consistency checks mitigate brittle failure modes common in direct prompting.

\subsubsection{Imputation}
Imputation is the main exception where \ours{} does not dominate. The best baseline (Zero-Shot+MM) achieves $6.729/10.758$ (MAE/MAPE), whereas \ours{} obtains $18.343/51.395$. This indicates that, for TSQA imputation, direct multimodal grounding can be more effective than multi-agent verification alone, likely because the task demands \emph{local, value-accurate} reconstruction rather than global structural reasoning. This result could indicate that local value reconstruction can be better served by direct visual grounding than cross-modal verification. A natural next step would be to integrate a value-calibrated imputation prior (or an explicit local interpolation module) within the \ours{} loop, using the verifier to validate rather than implicitly infer missing segments.

\subsubsection{Classification}
For classical classification, \ours{} attains the best accuracy ($0.607$), improving over the strongest baseline ($0.553$) by $+5.4$ points. This improvement is consistent with the advantage of cross-view verification for reducing spurious correlations and enforcing globally coherent rationales before committing to a discrete label.

\subsubsection{Anomaly Detection}
On anomaly detection, \ours{} achieves the best performance ($0.623$ ACC), surpassing the best baseline ($0.583$) by $+4.0$ points. These gains suggest that \ours{} is better at reconciling local deviations with overall temporal context, where purely pattern-matching strategies may overfit to short, noisy segments.

\subsubsection{Question Answering}
\ours{} performs strongly across QA formats: it achieves the best judgment accuracy ($0.897$) and multiple-choice accuracy ($0.796$), with the latter improving over the strongest baseline ($0.631$) by $+16.5$ points. For open-ended QA (NLI), \ours{} is close to the best baseline ($0.899$ vs.\ $0.911$), indicating that while verification improves factual consistency and decision accuracy, further gains on free-form generation may require stronger alignment between intermediate verification steps and the final natural-language output.

\subsection{Error Analysis: When Debate Helps and When It Does Not} \label{app:error_analysis}
Across the per-task results above, a consistent pattern emerges: \ours{} is most effective when the target depends on global structure, cross-view consistency, or relative quantities (trend classification, indicator differences, MCQA, anomaly detection, forecasting under structural regularity). It is less effective on tasks dominated by precise local value reconstruction\,(TSQA imputation, MTBench-Weather absolute extrema), where direct multimodal prompting can match or exceed it. We hypothesize that local reconstruction benefits less from cross-modal debate because there is little cross-modal disagreement to resolve---a single accurate numerical view suffices. Long-horizon causal reasoning and highly noisy multivariate inputs are plausibly affected by the same mechanism but were not isolated in our current evaluation; we mark this as a scope condition rather than a verified empirical claim. Integrating a value-calibrated interpolation prior within the VCC loop is a natural direction we leave to future work.

\begin{table*}[t]
\centering
\resizebox{\textwidth}{!}{%
\begin{tabular}{@{}c|ccccc|cccc|cccc|cc@{}}
\toprule
\multirow{2}{*}{\textbf{Backbone}} & \multicolumn{5}{c|}{\textbf{Classification (Accuracy) $\uparrow$}} & \multicolumn{4}{c|}{\textbf{Regression (MAE) $\downarrow$}} & \multicolumn{4}{c|}{\textbf{Question Answering (Accuracy) $\uparrow$}} & \multicolumn{2}{c}{\textbf{Computation Cost / Sample}} \\
 & \textbf{\begin{tabular}[c]{@{}c@{}}MTBench \\ (Finance)\end{tabular}} & \textbf{\begin{tabular}[c]{@{}c@{}}MTBench \\ (Weather)\end{tabular}} & \textbf{TimerBed} & \textbf{TSQA} & \textbf{Average} & \textbf{\begin{tabular}[c]{@{}c@{}}MTBench \\ (Finance)\end{tabular}} & \textbf{\begin{tabular}[c]{@{}c@{}}MTBench \\ (Weather)\end{tabular}} & \textbf{TSQA} & \textbf{Average} & \textbf{\begin{tabular}[c]{@{}c@{}}MTBench \\ (Finance)\end{tabular}} & \textbf{\begin{tabular}[c]{@{}c@{}}MTBench \\ (Weather)\end{tabular}} & \textbf{TSQA} & \textbf{Average} & \textbf{Est. Cost (\$)} & \textbf{Time Used (s)} \\
 \midrule
\midrule
gpt-4.1-mini & 37.50 & 34.00 & 28.47 & 33.50 & 33.37 & 0.74 & 3.91 & 90.17 & \textbf{31.61} & 84.00 & 55.00 & 80.00 & \textbf{73.00} & 0.0013 & 8.90 \\
gemini-2.5-flash & 32.50 & 56.00 & 28.65 & 29.50 & \textbf{36.66} & 4.68 & 3.68 & 106.35 & 38.24 & 70.00 & 50.00 & 76.00 & 65.33 & 0.0054 & 6.20 \\
grok-4.1-fast & 26.00 & 47.00 & 26.62 & 36.50 & 34.03 & 1.30 & 4.06 & 126.18 & 43.85 & 69.00 & 49.00 & 69.00 & 62.33 & 0.0006 & 3.50 \\
\bottomrule
\end{tabular}%
}
\caption{Zero-shot performance and efficiency comparison of candidate backbone models.} \label{table:backbone_prelims}
\end{table*}

\subsection{Backbone Selection} \label{section:backbone_prelims}

We evaluate three API-accessible backbones (\texttt{gemini-2.5-flash}, \texttt{gpt-4.1-mini}, and \texttt{grok-4.1-fast}) under the identical setup as the main experiments, using the same task suite, data sampling, chart generation, hyperparameters, and fixed random seeds. Each backbone is evaluated once, ensuring that performance differences are attributable solely to model capability.

\paragraph{Selection Criterion} As reported in \autoref{table:backbone_prelims}, \texttt{gpt-4.1-mini} offers the best overall trade-off between average task performance, inference latency, and estimated API cost. It achieves strong results across classification, regression, and question answering, while remaining computationally efficient. We therefore adopt \texttt{gpt-4.1-mini} as the default backbone for all main experiments.

\paragraph{Rationale for Non-Reasoning Backbones} Our objective is to evaluate whether \ours{} provides an explicit and portable reasoning scaffold for time-series understanding, rather than leveraging backbone-specific internal reasoning mechanisms. Using specialized reasoning models would obscure attribution by conflating protocol-level gains with proprietary chain-of-thought behaviors. We thus focus on fast, general-purpose backbones to ensure that observed improvements stem from the proposed debate and verification framework.

\begin{table}[t]
\centering
\resizebox{\linewidth}{!}{%
\begin{tabular}{@{}llccccc@{}}
\toprule
\textbf{Section} & \textbf{Benchmark} & $n$ & \textbf{\ours{}} & \textbf{Best Baseline} & \textbf{b--c / w--l} & $p$ \\ \midrule
Classification & MTBench (Fin) & 600   & 0.543 & 0.445 (BME)    & 152--93  & \textbf{2e-4} \\
Classification & MTBench (Wea) & 299   & 0.555 & 0.522 (ZS+MM)  & 67--57   & 0.42 \\
Classification & TimerBed      & 1{,}439 & 0.432 & 0.354 (CoT+MM) & 224--113 & \textbf{1.5e-9} \\
Classification & TSQA          & 593   & 0.614 & 0.560 (MAD)    & 128--96  & \textbf{0.038} \\
Classification & \textit{Average} & 2{,}931 & 0.504 & 0.406 (CoT+MM) & 553--265 & \textbf{3.6e-24} \\ \midrule
Regression & MTBench (Fin) & 553   & 0.807 & 0.993 (ZS+MM) & 274--279 & 0.86 \\
Regression & MTBench (Wea) & 598   & 3.584 & 3.710 (CoT)   & 306--270 & 0.14 \\
Regression & TSQA          & 600   & 27.8  & 47.5 (ZS+MM)  & 270--330 & 0.016$^{*}$ \\
Regression & \textit{Average} & 1{,}751 & 10.7 & 17.4 (ZS+MM) & 847--888 & 0.34 \\ \midrule
QA & MTBench (Fin)      & 300 & 0.913 & 0.847 (ZS+MM) & 27--7   & \textbf{8.2e-4} \\
QA & MTBench (Wea)      & 300 & 0.717 & 0.653 (ZS+MM) & 56--37  & 0.061 \\
QA & TSQA (exact-match) & 131 & 0.832 & 0.702 (MAD)   & 24--7   & \textbf{0.0033} \\
QA & \textit{Average}   & 731 & 0.818 & 0.718 (ZS+MM) & 124--51 & \textbf{3.4e-8} \\ \bottomrule
\end{tabular}%
}
\caption{Paired significance for each \autoref{table:main_results} benchmark$\times$task-type cell against its strongest baseline (accuracy: pooled exact McNemar, b--c = discordant pairs; regression: instance-level sign test, w--l = wins--losses; bold: $p<0.05$). Values are micro-averaged over paired instances and may differ slightly from \autoref{table:main_results}'s task averages (e.g., classification: 0.504 vs.\ 0.534). QA uses exact-match instances only, keeping these tests independent of the NLI proxy audited in \S\ref{app:scorer_validation}. $^{*}$For TSQA regression, the sign test favors the baseline (330 vs.\ 270) despite \ours{} nearly halving MAE because the test ignores error magnitude; the MAE gain is driven by avoiding large errors.} \label{tab:aggregate_significance}
\end{table}

\subsection{Robustness of the Aggregate Tests} \label{app:significance}
Because run subsets partially overlap, we (i) deduplicate repeated instances across runs (classification $p{=}9.0\times10^{-20}$; QA $p{=}4.5\times10^{-7}$; regression not significant), (ii) repeat the tests within each run (classification and QA remain significant throughout), and (iii) compare section averages with all eight baselines using Benjamini--Hochberg correction (classification and QA remain significant against every baseline for both pooled and deduplicated data; regression only against MAD+MM, ByMyEyes, and VL-Time). All significant cells in \autoref{tab:aggregate_significance} survive these checks.

\section{Resource Cost Comparison} \label{section:cost_comparison}

\begin{table}[t]
\centering
\resizebox{0.5\textwidth}{!}{%
\begin{tabular}{@{}lcccc@{}}
\toprule
\textbf{Method} & \textbf{Time (s)} & \textbf{\# Input Tokens} & \textbf{\# Output Tokens} & \textbf{Est. Cost (\$)} \\ \midrule
\midrule
Zero-Shot & 16.80 & 2356  & 234  & 0.001 \\
+ MM      & 14.30 & 6152  & 163  & 0.003 \\
CoT       & 15.70 & 2361  & 698  & 0.002 \\
+ MM      & 10.00 & 6157  & 416  & 0.003 \\
MAD       & 23.40 & 11460 & 1405 & 0.007 \\
+ MM      & 25.00 & 26499 & 1242 & 0.013 \\ \midrule
ByMyEyes  & 8.40  & 4565  & 379  & 0.002 \\
VL-Time   & 12.90 & 3900  & 758  & 0.003 \\ \midrule
\ours{} (Ours) & 70.70 & 68945 & 2883 & 0.033 \\ \bottomrule
\end{tabular}%
}
\caption{We report the average cost when using \texttt{gpt-4.1-mini} (\$0.40/1M input tokens and \$1.60/1M output tokens)} \label{table:cost_comparison}
\end{table}

We analyze LLM usage cost along wall-clock time, token consumption, and estimated monetary expense under a fixed backbone (\texttt{gpt-4.1-mini}) and identical evaluation settings. As shown in \autoref{table:cost_comparison}, \ours{} incurs higher cost than single-agent and lightweight multi-agent baselines, primarily due to deliberate reviewer verification, cross-modal evidence aggregation, and synthesis, which target numerical errors and modal inconsistencies. This additional overhead yields consistent performance gains in final accuracy (\autoref{table:main_results}) and remains modest in absolute monetary terms, making it a reasonable trade-off for error-sensitive applications. Moreover, \ours{} exposes practical cost-accuracy controls---such as reducing modalities, debate rounds, or reviewer count---that can substantially lower token usage while retaining much of the performance benefit (\S\ref{section:ablation_results}).

\ours{} is intended for offline or error-sensitive analysis rather than real-time, high-throughput forecasting. Its additional cost comes from deliberate verification and conflict resolution, which are most useful when users need auditable reasoning over numerical, visual, and textual evidence. For latency-sensitive settings, the same framework exposes cost controls through the number of reviewers, debate rounds, and enabled modalities; our sensitivity analysis shows that moderate configurations already capture most of the benefit. We therefore view \ours{} as a reliability-oriented inference protocol, not a low-latency replacement for direct prompting.

\section{Full Prompts for \ours{}} \label{section:prompts}
This section provides the \emph{system prompt} design for agent specifications and stage-wise \emph{instruction} for each agent.

\subsection{Agent Specifications} \label{section:agent_specs}

\subsubsection{Knowledge Elicitor}

\begin{promptBox}{Knowledge Elicitor}
\begin{lstlisting}
You are the world's best domain expert consultant for time-series analysis and reasoning.
You share specialized knowledge to help others solve these tasks correctly.

When given a task, share your expert knowledge about this domain:
- What do experts in this field know that helps solve such tasks?
- What are the typical ranges, patterns, and constraints?
- What mistakes do non-experts commonly make?

Be specific with numbers and concrete examples, while ensuring the knowledge is actionable by an LLM agent.
Do NOT solve the task or analyze data - only share domain expertise.
\end{lstlisting}
\end{promptBox}

\subsubsection{Numerical Analyst Agent}
\begin{promptBox}{Numerical Analyst}
\begin{lstlisting}
You are a data analyst for time-series tasks.
You work with raw numbers, statistics, and quantitative analysis of time-series data.

YOUR EXPERTISE (Data Analysis):
You analyze raw time series values, statistics, and computed features.

STRENGTHS:
- Exact values and precise calculations
- Statistical measures (mean, variance, trends)
- Quantitative comparisons

LIMITATIONS:
- Cannot access textual context or sentiment
- Cannot see visual patterns beyond what numbers show
- Calculations describe the PAST; extrapolating to FUTURE is inference, not fact
- Past momentum does not guarantee future continuation
- Can detect anomalies statistically, but their meaning (turning point? noise? error?) requires interpretation

{temporal_basics}

TOOLS (call `get_info` first):
- get_info() → schema, stats, detected features
- get_values(start, end) → time-series values by index or timestamp
- get_around(center, window) → time-series values around a point
- get_features(type) → 'peak'/'valley'/'trend'/'anomaly'
- get_frequency_features() → spectral analysis
- get_channel_values(ch, start, end) → specific channel
- get_all_channels(start, end) → all channels at once
- get_indicator(start, end) → indicator values (e.g., MACD, Bollinger Bands) of time series

**TOOL USAGE RULES** (STRICT):
1. THINK FIRST: Plan what you need before calling any tool.
2. MAX 5 CALLS TOTAL: After 5 calls, you MUST provide your evidence.
3. NO REPEATED CALLS: Never call the same tool with overlapping ranges.
4. EFFICIENT STRATEGY:
   - Call get_info() ONCE to understand the data
   - Use get_features() to find key points (peaks, anomalies)
   - Use get_around() for specific locations, NOT get_values() for large ranges
5. LEVERAGE HINTS: Use location hints from other analysts' observations when available.

{evidence_rules}
\end{lstlisting}
\end{promptBox}

\subsubsection{Text Analyst Agent}
\begin{promptBox}{Text Analyst}
\begin{lstlisting}
You are a text analyst for time-series tasks.
You read and interpret written content (reports, descriptions, contextual information) that provides context for time-series data.

YOUR EXPERTISE (Text Analysis):
You analyze reports, descriptions, and textual context about time series.

STRENGTHS:
- Contextual information (events, announcements, explanations)
- Sentiment and tone analysis
- Forward-looking statements that indicate potential changes
- Understanding WHY patterns might change (turning points, catalysts)

LIMITATIONS:
- Cannot provide exact numerical values or precise calculations
- Cannot see chart patterns or visual trends
- Forward-looking statements (forecasts, analyst opinions) are interpretations, not verified facts
- Distinguish between reported events and speculative commentary

{temporal_basics}

{evidence_rules}
\end{lstlisting}
\end{promptBox}

\subsubsection{Visual Analyst Agent}
\begin{promptBox}{Visual Analyst}
\begin{lstlisting}
You are a chart analyst for time-series tasks.
You examine visual patterns in time-series charts and plots.

YOUR EXPERTISE (Chart Analysis):
You analyze time series charts, plots, and visual patterns.

STRENGTHS:
- Overall trend direction (upward, downward, stable)
- Pattern recognition (cycles, anomalies, breakouts)
- Shape of recent movement (stabilizing, accelerating, reversing)
- Comparative visual analysis

LIMITATIONS:
- Cannot read precise numerical values from charts
- Cannot access textual context
- What you SEE ("the chart shows X") differs from what you INFER ("this suggests Y")
- Pattern interpretation requires stating what is observed vs. what is concluded

{temporal_basics}

{evidence_rules}
\end{lstlisting}
\end{promptBox}

\subsubsection{Reviewer Agent}
\begin{promptBox}{Reviewer Agent Profile}
\begin{lstlisting}
{knowledge_section}

You are an expert reviewer for a time-series reasoning task.
You evaluate evidence from specialist analysts and synthesize a well-reasoned answer.

YOUR JOB:
- Score each analyst's evidence based on observation quality, inference logic, and honesty
- Verify claims against original data before accepting them
- Identify conflicts between different analysts
- Adjust your answer confidence based on how well evidence agrees
- Check claims against domain knowledge (if provided above)

AVAILABLE FOR FACT-CHECKING:
- VISUAL: time series + frequency charts attached
- NUMERICAL: get_info, get_values, get_around, get_features, get_frequency_features, get_channel_values, get_all_channels, get_indicator
- CODE: execute_code(code) for calculations

MANDATORY VERIFICATION:
Before accepting ANY claim, VERIFY against available sources:
- Available: numerical lookup tools (get_values, get_features), code executor for calculations, attached charts, text context (embedded in task description)
- Mark claims as VERIFIED, UNVERIFIED, or CONTRADICTED
- Lower weight for UNVERIFIED claims, reject CONTRADICTED claims

**TOOL USAGE RULES** (STRICT):
1. THINK FIRST: Identify which claims need verification before calling any tool.
2. MAX 3 CALLS TOTAL: After 3 calls, you MUST synthesize your answer.
3. NO REPEATED CALLS: Never call the same tool with overlapping ranges.
4. EFFICIENT STRATEGY:
   - Call get_info() ONCE to understand the data
   - Use get_features() to find key points, then get_around() for specific values
   - Use execute_code() for complex calculations
   - Leverage numerical data embedded in task description

{TEMPORAL_AWARENESS}

{JUDGE_EVALUATION_CRITERIA}

{JUDGE_PROTOCOL}
\end{lstlisting}
\end{promptBox}

\subsubsection{Final Synthesizer}
\begin{promptBox}{Final Synthesizer Profile}
\begin{lstlisting}
{knowledge_section}

You are the final decision-maker for a time-series reasoning task.
You determine the correct answer based on expert reviewers' evaluations.

YOUR JOB:
- Evaluate each reviewer's reasoning quality
- Check if reviewers followed the SUGGESTED APPROACH from DOMAIN KNOWLEDGE
- Verify disputed answers against data (if about past) or domain reasoning (if about future)
- Derive your own answer if all reviewers made the same error
- Always be skeptical of the reviewers' answers; do not blindly trust them

AVAILABLE FOR VERIFICATION:
- VISUAL: time series + frequency charts attached
- NUMERICAL: get_info, get_values, get_around, get_features, get_frequency_features, get_channel_values, get_all_channels, get_indicator
- CODE: execute_code(code) for calculations

MANDATORY VERIFICATION:
Before accepting ANY claim, VERIFY against available sources:
- Available: numerical lookup tools (get_values, get_features), code executor for calculations, attached charts, text context (embedded in task description)
- Mark claims as VERIFIED, UNVERIFIED, or CONTRADICTED
- Lower weight for UNVERIFIED claims, reject CONTRADICTED claims

**TOOL USAGE RULES** (STRICT):
1. THINK FIRST: Identify which claims need verification before calling any tool.
2. MAX 3 CALLS TOTAL: After 3 calls, you MUST synthesize your answer.
3. NO REPEATED CALLS: Never call the same tool with overlapping ranges.
4. EFFICIENT STRATEGY:
   - Call get_info() ONCE to understand the data
   - Use get_features() to find key points, then get_around() for specific values
   - Use execute_code() for complex calculations
   - Leverage numerical data embedded in task description

{TEMPORAL_AWARENESS}

{SYNTHESIZER_PROTOCOL}
\end{lstlisting}
\end{promptBox}

\subsection{Temporal Awareness for Agents} \label{section:temporal_prompts}

\subsubsection{Instruction for Debating Agents}
\begin{promptBox}{Basic Temporal Reasoning}
\begin{lstlisting}
TEMPORAL REASONING:
- Your data ends at time T (past data). You cannot see the future.
- Any prediction about future values is an INFERENCE from past patterns.

SIGNAL TYPES:
- Forward-looking signals (context about future conditions) inform predictions
- Historical trends show momentum but may miss turning points
- When forward vs backward signals CONFLICT: note this in your evidence
\end{lstlisting}
\end{promptBox}

\subsubsection{Basic Instruction for Reviewer Agents}
\begin{promptBox}{Temporal Awareness for VCC}
\begin{lstlisting}
TEMPORAL REASONING:
- Data ends at time T (past data). Future is unknown.
- Predictions about future are INFERENCES from past patterns.

KEY QUESTION FOR PREDICTIONS: Will the CAUSE of the observed trend persist?
- If cause persists → trend may continue
- If cause is ending/changing → trend may change or stabilize
- Causal/leading indicators outweigh lagging historical patterns
- New information can invalidate historical trends
- EXTRAPOLATION (bad): Assuming past trends continue blindly
- PREDICTION (good): Reasoning about whether conditions will persist

TASK TYPES:
- FUTURE: Answer about something that hasn't happened yet
  → Forward-looking signals are PRIMARY (what's changing?)
  → Historical trends are SECONDARY (show momentum, miss turning points)
  → Past data CANNOT verify future predictions - only domain knowledge-based reasoning can
  
- PAST-PRESENT: Answer about something that has happened
  → Data verification is PRIMARY - check claims against data
  → External context is SECONDARY

VERIFICATION RULES:
- PAST-PRESENT claims: VERIFY or CONTRADICT with data
- FUTURE claims: Mark as UNVERIFIABLE by data (not "contradicted")
  → Evaluate domain knowledge-based reasoning quality instead
\end{lstlisting}
\end{promptBox}

\subsection{Multimodal Query Construction} \label{section:mqc_prompts}

\subsubsection{Knowledge Elicitation}
\begin{promptBox}{Knowledge Elicitation Prompt}
\begin{lstlisting}
TASK DESCRIPTION:
{task_description}

Share domain knowledge to help an LLM agent solve this task correctly. Be SPECIFIC with numbers, ranges, and patterns.
DO NOT provide an answer - only domain knowledge and key insights.

1. DOMAIN: What domain and task type (classification/forecasting/imputation/anomaly_detection/QA/etc.)?

2. KNOWLEDGE & KEY SIGNALS: What domain knowledge and patterns are relevant?
   - Physical/domain constraints, typical ranges, expected behaviors
   - What patterns distinguish different outcomes?
   - What we need to know to answer THIS task correctly?

3. SUGGESTED APPROACH: How should data be analyzed by an LLM agent?
   - What features to examine? How to interpret them?
   - How to interpret observations correctly?
   - How to extract findings and make inferences from the data?

4. PITFALLS: Common mistakes to avoid?
   - What looks similar but means different things?
   - What domain knowledge is often overlooked?
   - How to avoid common pitfalls and errors?

5. MODALITY: Which modalities [TEXT, VISUAL, NUMERICAL, FREQUENCY] to focus on (important modalities)?
   - Which modalities are most likely to be relevant and decisive?
   - Which modalities are most likely to be misleading and harmful?
  

OUTPUT (only answer to the questions listed above, NOT to the task description):
DOMAIN: [domain and task type]
KNOWLEDGE: [relevant knowledge with numbers/ranges]
KEY SIGNALS: [patterns to look for]
SUGGESTED APPROACH: [how to analyze correctly]
PITFALLS: [common mistakes]
MODALITY: [modalities to focus on]
\end{lstlisting}
\end{promptBox}

\subsubsection{Knowledge Inclusion}
\begin{promptBox}{Knowledge Inclusion Prompt}
\begin{lstlisting}
DOMAIN KNOWLEDGE:

{domain_knowledge}

USE THIS KNOWLEDGE to guide your analysis:
- Verify observations against these expected patterns
- Flag observations that CONTRADICT domain expectations
- Apply domain constraints when making inferences
\end{lstlisting}
\end{promptBox}

\subsection{Multimodal Collaborative Debate} \label{section:debate_instructions}

\subsubsection{Evidence Templates for Debating Agents}
\begin{promptBox}{First Round Template}
\begin{lstlisting}
Task: {task}

RESPOND IN THIS EXACT FORMAT (MAX 150 WORDS):

UNDERSTANDING: <Restate the question in your own words. What is being asked? What type of answer is needed? 1-2 sentences>

USEFUL OBSERVATIONS:
1. <Specific observation from your {modality_name} data> [OBSERVATION]
2. <Specific observation from your {modality_name} data> [OBSERVATION]

INFERENCES:
1. <What do these observations suggest? Use domain knowledge. 1-2 sentences> [INFERENCE]

LIMITS: <What can {modality_name} NOT determine? Be honest. 1-2 sentences>

Stop here. Do NOT provide a final answer. The reviewer synthesizes the final answer.
\end{lstlisting}
\end{promptBox}

\begin{promptBox}{Subsequent Round Template}
\begin{lstlisting}
=== PREVIOUS ROUND EVIDENCE ===
{debate_history}
=== END PREVIOUS EVIDENCE ===

Task: {task}

RESPOND IN THIS EXACT FORMAT (MAX 150 WORDS):

UNDERSTANDING: <Restate the question in your own words. What is being asked? What type of answer is needed? 1-2 sentences>

OTHER PERSPECTIVES: <Summarize what other analysts reported. 1-2 sentences>

USEFUL OBSERVATIONS:
1. <Maintain your key observation - do NOT abandon it> [OBSERVATION]
2. <Another observation if applicable> [OBSERVATION]

INFERENCES:
1. <Given your observations AND others' evidence, what does this suggest? 1-2 sentences> [INFERENCE]

LIMITS: <What can {modality_name} still NOT determine? Be honest. 1-2 sentences>

Stop here. Do NOT provide a final answer. The reviewers decide.
\end{lstlisting}
\end{promptBox}

\subsubsection{Evidence Rules}
\begin{promptBox}{Evidence Writing Instruction}
\begin{lstlisting}
EVIDENCE RULES:

YOUR ROLE: Present evidence from your data source. A reviewer synthesizes the final answer.

REQUIREMENTS:
✓ Label clearly: [OBSERVATION] vs [INFERENCE]
✓ Be specific and verifiable - support claims with data or references
✓ Acknowledge limitations honestly
✗ Do NOT provide final answers
✗ Do NOT exaggerate or overclaim

IN REFINEMENT ROUNDS (Round 2+):
✓ PRESERVE your original observations
✓ Acknowledge other analysts' evidence
✓ EXPLAIN how your observations relate to theirs (support/contradict/add context)
✗ Do NOT abandon your position to match others
\end{lstlisting}
\end{promptBox}

\subsection{VCC Protocol}

\subsubsection{Scoring Instructions}
\begin{promptBox}{Scoring Rubric}
\begin{lstlisting}
SCORING CRITERIA (100 points max per analyst):

INFERENCE QUALITY (0-50 pts) - Most important
- Logic (0-25): Inferences follow from observations? Uses domain knowledge?
- Calibration (0-25): Appropriately cautious? Avoids blind extrapolation?

OBSERVATION QUALITY (0-30 pts)
- Specificity (0-15): Concrete, verifiable details from data?
- Labeling (0-10): Correctly distinguished [OBSERVATION] vs [INFERENCE]?
- Relevance (0-5): Observations address the question?

HONESTY (0-20 pts)
- Limits (0-10): Acknowledged what data cannot determine?
- No Overclaiming (0-10): Avoided exaggeration?
\end{lstlisting}
\end{promptBox}

\begin{promptBox}{Review Protocol}
\begin{lstlisting}
REVIEW PROTOCOL:

STEP 1 - SCORE ANALYSTS (0-100 each):
- Inference Quality (50): Logic + calibration
- Observation Quality (30): Specificity + labeling + relevance
- Honesty (20): Limits acknowledged + no overclaiming
- Weight = score / sum(scores). Reject if score < 40.

STEP 2 - VERIFY CLAIMS:
Check each major claim against data AND domain knowledge:
- DATA: VERIFIED / UNVERIFIED / CONTRADICTED
- DOMAIN: MATCHES / VIOLATES / N-A
- REJECT claims that are CONTRADICTED or VIOLATE domain

STEP 3 - DETECT CONFLICTS:
- DIRECT: Analysts give opposite conclusions
- PARTIAL: Disagree on details
- NO CONFLICT: Agree or address different aspects

STEP 4 - SYNTHESIZE ANSWER:
Match confidence to evidence agreement:
- NO CONFLICT + VERIFIED → confident answer
- CONFLICT UNRESOLVED → MODERATE/CONSERVATIVE answer
- DOMAIN VIOLATION → reject that claim, choose domain-consistent answer
- FUTURE task → evaluate reasoning quality (data can't verify predictions)
\end{lstlisting}
\end{promptBox}

\begin{promptBox}{Reviewer Prompt}
\begin{lstlisting}
Task: {task}

Evidence:
{final_debate_history}

ANALYSTS: TEXT, VISUAL, NUMERICAL

YOUR JOB: Score evidence, VERIFY claims (data + domain), detect CONFLICTS, synthesize CALIBRATED answer.

OUTPUT FORMAT (keep everything concise):
TASK: <Restate what the task is asking in one sentence>
TASK TYPE: [FUTURE / PAST-PRESENT]

SCORES:
- TEXT: (Observation: _/30, Inference: _/50, Honesty: _/20) = [0-100]
- VISUAL: (Observation: _/30, Inference: _/50, Honesty: _/20) = [0-100]
- NUMERICAL: (Observation: _/30, Inference: _/50, Honesty: _/20) = [0-100]

WEIGHTS:
- TEXT: [X%]
- VISUAL: [X%]
- NUMERICAL: [X%]

VERIFICATION (check against lookup tools, code executor, charts, text in task, domain knowledge (above)):
- [Claim]: [VERIFIED/UNVERIFIED/CONTRADICTED] + [DOMAIN: MATCHES/VIOLATES/N-A] - [explanation]
- [Claim]: [VERIFIED/UNVERIFIED/CONTRADICTED] + [DOMAIN: MATCHES/VIOLATES/N-A] - [explanation]
(For each major claim: first check domain knowledge, then check data sources)

**IMPORTANT**: Never verify claims against data for FUTURE tasks. Past time series CANNOT verify future predictions. Always check the requested date against the date at which the data actually ends.
Do NOT refer to past data calculation or verification to support the answer. Only past data vs. future data (expected values by YOUR prediction) can be used to support the answer.

OUTSTANDING CONFLICTS: [NO_CONFLICT / DETECTED / RESOLVED] - <details>
KEY EVIDENCE: <main observations/inferences used>
CALIBRATED ANSWER: [answer in exact task required format]
\end{lstlisting}
\end{promptBox}

\subsubsection{Synthesizer Instructions}
\begin{promptBox}{Decision Protocol}
\begin{lstlisting}
DECISION PROTOCOL:

STEP 1 - CHECK APPROACH (Critical First):
- What did DOMAIN KNOWLEDGE say to do?
- What did reviewers actually do?
- If MISMATCH: Reviewers answered the WRONG question → derive your own answer

STEP 2 - SCORE REVIEWERS (0-100 each):
- Task Understanding (20): Did they follow suggested approach? (0 if mismatch!)
- Evidence Usage (20): Used evidence correctly?
- Verification (20): Verified claims before accepting?
- Conflict Handling (20): Detected and adjusted for conflicts?
- Calibration (20): Confidence matches evidence?

STEP 3 - IDENTIFY TASK TYPE:
- FUTURE task: Past data CANNOT verify predictions
  → Check for BLIND EXTRAPOLATION: Did they ask "will the CAUSE persist?"
  → Evaluate domain reasoning quality
- PAST-PRESENT task: Data CAN verify claims

STEP 4 - DERIVE ANSWER:
- UNANIMOUS + CORRECT APPROACH → use shared answer
- UNANIMOUS + WRONG APPROACH → REJECT, derive your own
- SPLIT + UNRESOLVED → CONSERVATIVE answer
- DOMAIN VIOLATION → reject that answer

CRITICAL RULES:
✗ Don't trust unanimous if wrong approach
✗ Don't pick majority without verification
✗ Don't accept answers that violate domain knowledge
✓ Check approach FIRST
✓ Always be skeptical of the reviewers' answers; do not blindly trust them
✓ Be conservative when uncertain
\end{lstlisting}
\end{promptBox}

\begin{promptBox}{Synthesizer Prompt}
\begin{lstlisting}
Task: {task_description_judge}

ANALYSTS: TEXT, VISUAL, NUMERICAL

Reviewer Evaluations:
{all_judge_responses}

---

VERIFICATION SOURCES (max 3 tool calls when available):
- NUMERICAL TOOLS: get_info, get_values, get_around, get_features, get_frequency_features, get_channel_values, get_all_channels, get_indicator
- CODE: execute_code(code)
- CHARTS: time series, frequency (attached)
- TEXT: embedded in task description
- DOMAIN: knowledge in system prompt

OUTPUT FORMAT (keep everything concise):
TASK: <Restate what the task is asking in one sentence>
TASK TYPE: [FUTURE / PAST-PRESENT]

APPROACH CHECK:
- SUGGESTED: <from domain knowledge>
- USED: <by reviewers>
- Status: <always re-check even with UNANIMOUS answers> [CORRECT / MISMATCH]

REVIEWER SCORES:
- Reviewer 0: (Task: _/20, Evidence: _/20, Verification: _/20, Conflicts: _/20, Calibration: _/20) = [0-100]
- Reviewer 1: (Task: _/20, Evidence: _/20, Verification: _/20, Conflicts: _/20, Calibration: _/20) = [0-100]
Note that perfect score is IMPOSSIBLE. Never give 100 score to any reviewer. There are always flaws in the reviewers' reasoning.

ANSWER VERIFICATION (use TASK TYPE from above, summarizing only outstanding concerns):
- If PAST-PRESENT task: Verify answers against data
- If FUTURE task: Past data describes history but CANNOT verify predictions - evaluate domain reasoning

**IMPORTANT**: Never verify claims against data for FUTURE tasks. Past time series CANNOT verify future predictions. Always check the requested date against the date at which the data actually ends. 
Do NOT refer to past data calculation or verification to support the answer. Only past data vs. future data (expected values by YOUR prediction) can be used to support the answer.

CONFLICT STATUS:
- Reviewer Agreement: [UNANIMOUS / SPLIT / ALL_DIFFERENT]
- Approach Status: [ALL_CORRECT / ALL_WRONG / MIXED]
- Analyst Agreement: [From reviewer reports - did analysts conflict?]
- Resolution: [VERIFIED_RESOLUTION / UNRESOLVED / NO_CONFLICT / APPROACH_ERROR]

CALIBRATED REASONING (apply based on TASK TYPE, summarizing in 1-2 short sentences):
- FUTURE task: <evaluate domain reasoning quality - cannot verify with past data; if reviewers said verified for FUTURE task, they are wrong!>
- PAST-PRESENT task: <data verification result>
- If MISMATCH: <derive own answer using correct approach>
- If SPLIT + cannot resolve: <Choose CONSERVATIVE answer>

FINAL ANSWER: <your calibrated answer in exact, user-defined task format>
\end{lstlisting}
\end{promptBox}

\section{Case Studies from Stored Traces} \label{app:case_studies}
We present two samples verbatim from the stored traces: one in which verification detects and resolves a cross-modal conflict, and one that marks the boundary of what verification can arbitrate.

\subsection{Cross-modal conflict detected and resolved by verification} 
MTBench weather trend (next-day mean-temperature trend, threshold $\pm$0.5\,\textdegree C), main run 1, sample 13; ground truth ``decreasing''.
\begin{promptBox}{Case Study 1: Success}
\begin{lstlisting}
- Visual analyst (rounds 1 and 2): "the mean temperature for the last
  24 hours likely does not differ from the next day's expected mean by
  more than +-0.5C, suggesting a stable temperature trend."
- Reviewer-agent verification (numerical lookup): "The temperature
  trend for the next day is stable or slightly cooling based on
  observations. [CONTRADICTED by numerical data] - Numerical data
  shows a significant cooling trend beyond the +-0.5C threshold. Key
  evidence: "July 8 mean = 22.75C, July 7 mean = 24.05C, difference =
  -1.3C (cooling)." Re-deriving the daily means from the raw values
  gives 24.046 -> 22.750 (-1.296C): the reviewer agent's lookup is
  exact.
- Conflict record: DETECTED ("Text and visual analysts inferred stable
  trend ... but numerical analyst shows a clear cooling trend");
  Resolution: VERIFIED_RESOLUTION.
- Final answer: "decreasing" (correct). Persuasion- or vote-based
  resolution would have favored the two agreeing (wrong) analysts.
\end{lstlisting}
\end{promptBox}

\subsection{Shared wrong prior leaves nothing to arbitrate} 
Same task, sample 16; ground truth ``decreasing''.
\begin{promptBox}{Case Study 2: Failure}
\begin{lstlisting}
- All three analysts converge on warming. Text: "The arrival of mild,
  humid southwesterly flow after the storm likely leads to a warming
  trend or at least prevents a temperature decrease." Numerical: "The
  temperature increased by about 6F from March 30 to March 31,
  indicating a strong warming trend likely to continue into the next
  day." Visual: "The increasing temperature pattern and stable diurnal
  cycle suggest a continuing warming trend."
- Reviewer-agent records: "OUTSTANDING CONFLICTS: NO_CONFLICT - All
  analysts agree on warming trend". The forward-looking claim is
  correctly labeled "UNVERIFIED (future prediction) + DOMAIN: MATCHES".
- Final answer: "increasing", which is wrong.
\end{verbatim}
\end{lstlisting}
\end{promptBox}
The observed warming is real (daily means 5.8\,\textdegree F on Mar~30 $\to$ 13.3\,\textdegree F on Mar~31), so no past--present check can falsify the extrapolation, and with every modality sharing the same plausible prior there is no disagreement for VCC to arbitrate. This is the boundary stated in \S\ref{section:vcc_protocol} and \S\ref{app:error_analysis}: future-oriented claims can only be calibrated for domain plausibility, not verified against data.

\section{Use of Large Language Models}
In preparing this submission, we used ChatGPT for language refinement, including improving clarity and correcting grammatical and typographical errors. Its use was limited to copyediting tasks such as sentence restructuring and rephrasing for readability. All LLM-assisted edits were carefully reviewed and, where necessary, revised by the authors to ensure accuracy, originality, and full compliance with research-integrity standards. We note that large language models also serve as the backbone for \emph{all} methods evaluated in our experiments, as described above.

\end{document}